\documentclass[11pt,a4paper]{article}
\usepackage[hyperref]{acl2021}
\usepackage{times}
\usepackage{latexsym}

\usepackage{microtype}

\usepackage{microtype}
\usepackage{graphicx}
\usepackage{paralist}
\usepackage{enumitem}
\usepackage{siunitx}
\usepackage{subcaption}
\usepackage{xcolor}
\usepackage{lipsum}
\usepackage{multirow}
\usepackage{color}
\usepackage{wrapfig}
\usepackage{amsmath}
\usepackage{subcaption}

\title{Towards Navigation by Reasoning over Spatial Configurations}

\author{Yue Zhang \\
  Michigan State University\\
  \texttt{zhan1624@msu.edu} \\\And
  Quan Guo \\
  Michigan State University\\
  \texttt{guoquan@msu.edu} \\ \And
  Parisa Kordjamshidi\\
  Michigan State University\\
  \texttt{kordjams@msu.edu}}

\date{}

\begin{document}
\maketitle
\begin{abstract}
We deal with the navigation problem where the agent follows natural language instructions while observing the environment.
Focusing on language understanding, we show the importance of spatial semantics in grounding navigation instructions into visual perceptions. 
We propose a neural agent that uses the elements of spatial configurations and investigate their influence on the navigation agent's reasoning ability. Moreover, we model the sequential execution order and align visual objects with spatial
configurations in the instruction. Our neural agent improves strong baselines on the seen environments and shows competitive performance on the unseen environments. Additionally, the experimental results demonstrate that explicit modeling of spatial semantic elements in the instructions can improve the grounding and spatial reasoning  of the model.
\end{abstract}

\section{Introduction}
The ability to understand and follow natural language instructions is critical for intelligent agents to interact with humans and the physical world.
One of the recently designed tasks in this direction is Vision-and-Language Navigation~(VLN)~\cite{anderson2018vision}, which requires an agent to carry out a sequence of actions in a photo-realistic simulated environment in response to a sequence of natural language instructions.
To accomplish this task, the agent should have three abilities: understanding linguistic semantics, perceiving the visual environment, and reasoning over both modalities~\cite{zhu2020babywalk,wang2019reinforced}.
While understanding vision and language are difficult problems by themselves, learning the connection between them without direct supervision makes this task even more challenging~\cite{hong2020sub}. 
 
To address this challenge, some neural agents establish the connection using attention mechanism to relate the tokens from a given instruction to the images in a panoramic photo~\cite{anderson2018vision,fried2018speaker,ma2019self,yu2018guided}. 
Surprisingly, although those models can improve the performance,~\newcite{hu2019you} found they ignore the visual information.
There is no clear evidence that the agent can correspond the components of the visual environment to the instructions~\cite{hong2020sub}.
Based on these results, recent research started to improve the agent's reasoning ability by explicitly considering the structure of language and image.
From the language side, \newcite{hong2020sub} annotated fine-grained sub-instructions and their corresponding trajectories and used the co-grounded features of a part of instruction and the image to predict the next action. 
From the image side, \newcite{hu2019you} induced a high-level object-based visual representation to ground the language into the visual context.

In the same direction, we propose a neural agent, namely \textit{Spatial-Configuration-Based-Navigation (SpC-NAV)}, and consider the structure of both modalities, that is, spatial semantics of the instructions  and the objects in the images. 
We use the notion of  \textit{Spatial Configuration}~\cite{dan2020spatial} to model the instructions and design a state attention  to ensure the execution order of spatial configurations.
Then, we utilize the spatial semantics elements, namely \textit{motion indicator}, \textit{spatial indicator} and \textit{landmark} in spatial configuration to establish the connection with the visual environment. Specifically, we use the similarity score between the landmark representation in the spatial configurations and the object representation in the panoramic images to control the transitions between configurations.
Also, we align object representations with the configuration representations enriched with motion indicator, spatial indicator and landmark representations to finally select the navigable image.



\begin{figure}[!ht]
\centering
\begin{subfigure}[b]{0.90\linewidth}
\includegraphics[width=\linewidth]{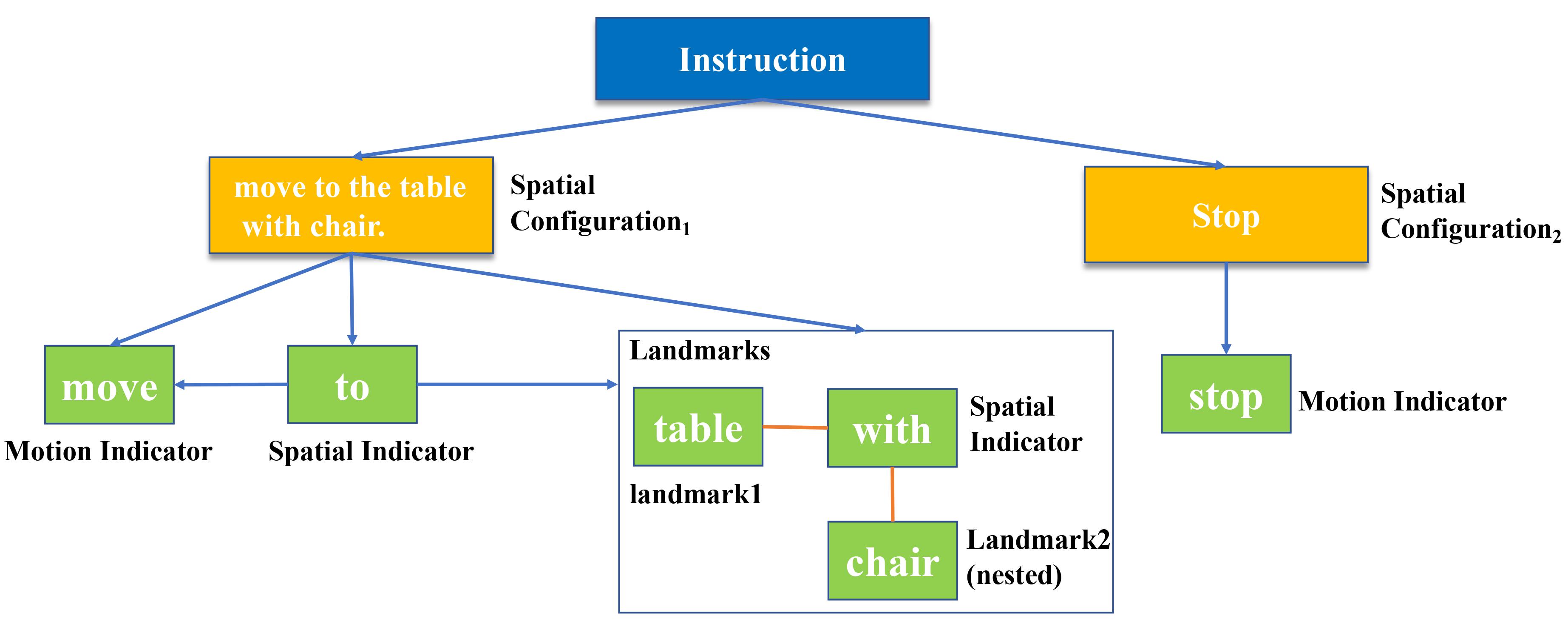}
\caption{Spatial Configuration Scheme}
\label{figure1}
\end{subfigure}
\begin{subfigure}[b]{1.00\linewidth}
\includegraphics[width=\linewidth]{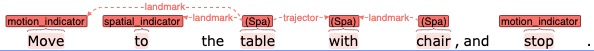}
\caption{Spatial Configuration Annotation}
\label{figure2}
\end{subfigure}
 \caption{\textbf{Spatial Configuration example}. The instruction "Move to the table with chair, and stop." can be split into two spatial configurations: "move to the table with chair" and "stop". In configuration1, "move" is motion indicator; "to" is spatial indicator; "table" is landmark. "table with chair" is a nested spatial configuration of configuration1. The role of "table" is  trajector; "with" is spatial indicator; and "chair" is landmark. In configuration2, "stop" is motion indicator.}
 \vspace{-5mm}
\label{fig:spatial Configuration Scheme Example}
\end{figure}
A spatial configuration is the smallest linguistic unit that describes the location/trans-location of an object with respect to a reference or a path that can be perceived in the environment. It contains fine-grained spatial roles, such as motion indicator, landmark, spatial indicator, trajector. Essentially, each spatial configuration forms a sub-instruction in our setting. 
Figure \ref{fig:spatial Configuration Scheme Example} shows an example of splitting an instruction into its corresponding spatial configurations and the extracted spatial roles. 
Previous research argues representing the semantic structure of the language could improve the reasoning capabilities of deep learning models~\cite{dan2020spatial, zheng2020srlgrn}.
There are relevant work modeling the meaning of spatial semantics in probabilistic models~\cite{kollar2010toward,tellex2011understanding} and neural models~\cite{regier1996human, ghanimifard2019goes}. However, its impact on deep learning models for navigation remains an open research problem.

\noindent The contribution of this paper is as follows:\\
1. We consider the spatial semantic structure of the instructions explicitly in terms of spatial configurations and their spatial semantic elements, i.e., spatial/motion indicators, and landmarks to enrich the configuration representations.\\
2. We introduce a state attention to guarantee that configurations are executed sequentially. Also, we utilize the grounding between the extracted spatial elements and the object representation to help control the transitions between configurations.\\
3. Our experiment results show that considering the explicit representation of semantic elements of the spatial configurations improves the strong baselines significantly in the seen environments and yields competitive results in the unseen environments. 

\section{Related Work}
\label{Related Work}
Older studies on navigation before the deep learning era are mostly symbolic grounding methods, which are based on parsing the semantics of the instruction and learning probabilistic models. \newcite{macmahon2006walk} used the parser to associate the linguistic elements in free-form instruction to their corresponding action, location and object in the environment. 
\newcite{tellex2011understanding} represented the spatial language as a hierarchy of Spatial Description Clauses (SDC) and proposed a discriminative probabilistic graphical model to find the most probable path with the extracted SDC and the detected visual landmark.
\newcite{mei2016listen} provided a good overview of the past classical work on navigation. However, one of the biggest limitations of those methods is that they required prior linguistic structure and manual annotations.

In recent years, given the new capabilities created by deep learning architectures, the navigation task is extended to the photo-realistic simulated environments~\cite{anderson2018vision,thomason2019vision,chen2019touchdown}. 
Based on this,
a Sequence-to-Sequence (Seq2seq) baseline model was proposed by~\newcite{anderson2018vision} to encode the instructions and decode the embeddings to identify the corresponding output action sequence with the observed images. \newcite{fried2018speaker} proposed to train a speaker model to augment the instructions for the follower model. \newcite{ma2019self} introduced a visual and textual co-attention mechanism and a progress monitor loss to track the execution progress.
%
Although those agents achieved better performance, 
the semantic structures on both language and vision sides were ignored. 

We aim to exploit both symbolic grounding and neural models in the spatial domain.  \newcite{regier1996human} designed the neurons to learn the meaning of spatial prepositions.
\newcite{ghanimifard2019goes} explored the effects of spatial knowledge in a generative neural language model for the image description. We mainly work on incorporating the spatial semantics in navigation neural agent.
\newcite{hong2020sub} recently provided a method to segment the long instruction into sub-instructions. They used a shifting attention module to infer whether the current sub-instruction has been completed. Sub-instructions differ from us as 
they manually aligned the instructions and viewpoints to learn the alignments, while we modeled spatial semantics to guide the alignment automatically.
Moreover, their proposed shifting attention module is hard attention, and a threshold is set to decide whether the agent should execute the next sub-instruction. However, we utilize the grounding between the landmarks and the objects to control the transitions between sub-instructions. 
\
\section{Navigation Model}
\subsection{Problem Formulation}
In this task, the agent follows an instruction to navigate from a start viewpoint to a goal viewpoint in a photo-realistic environment. Formally, the agent is given a natural language instruction $S$, which is a sequence of tokens, and $\{s_{1}, s_{2}, \cdots\}$ is its corresponding token embeddings. The agent observes a 360-degree panoramic view of its surrounding scene at the current viewpoint.
Here, we follow \newcite{ma2019self} to map the $n$ navigable viewpoints to discrete images from the current panoramic view\footnote{12 headings and 3 elevations with 30 degree interval.}. We obtain $n$ images corresponding to each navigable viewpoint $I=\{I_{1},I_{2},\cdots,I_{n}\}$.
The task is to select the next viewpoint among the navigable viewpoints or the current viewpoint (indicating the stop), and finally, to generate the trajectory that takes the agent close to an intended goal location.

\begin{figure*}[!t]
    \centering
    \includegraphics[width=4.0in]{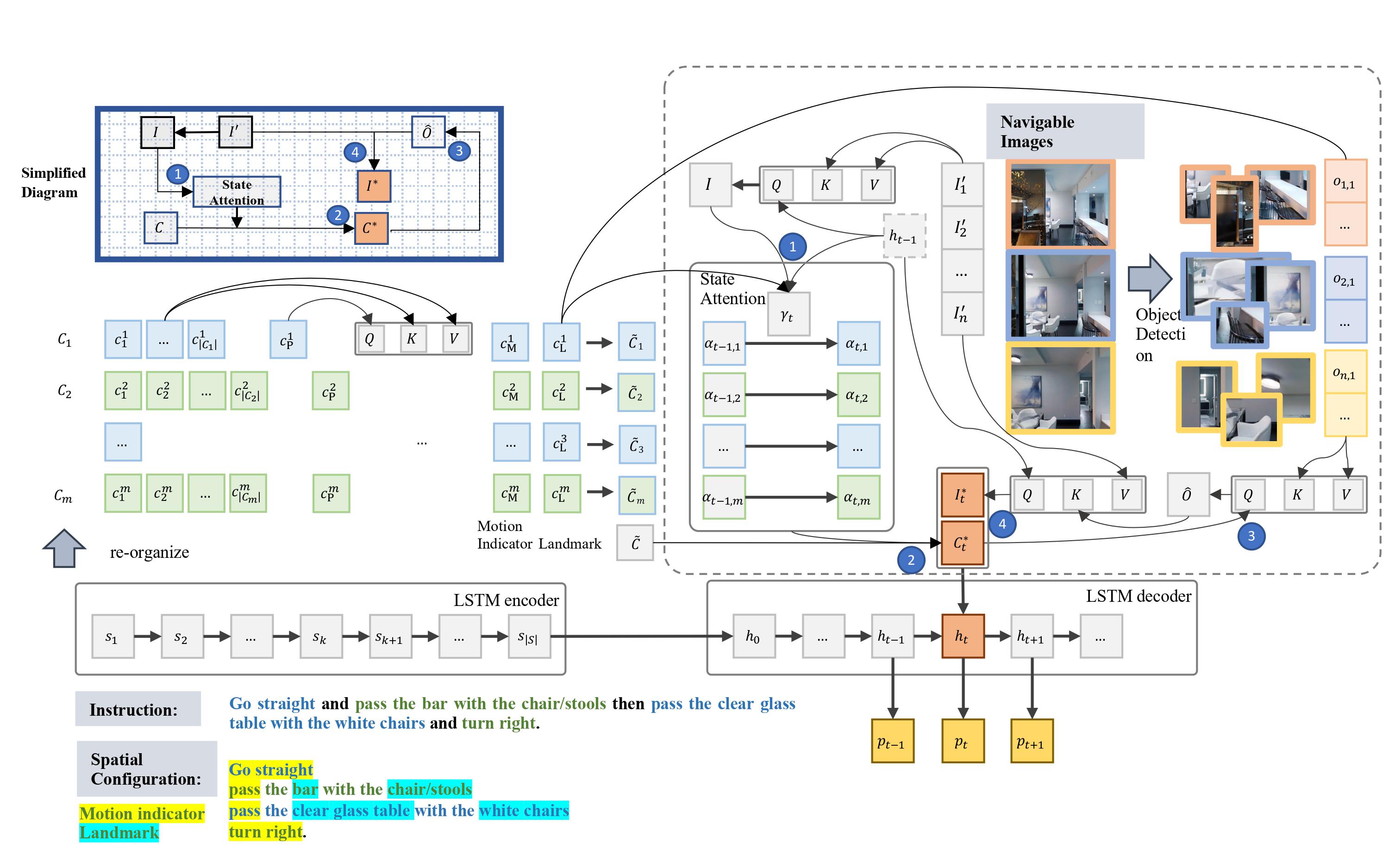}
    \caption{\textbf{Model Architecture.} 
    The input to the encoder is the instruction text. The inputs to the decoder are the grounded language $C^*_t$ calculated by state attention and the aligned visual representations $I^*_t$ obtained from navigable images at each step $t$. The decoder predicts the distribution of next viewpoint $p_t$ with the updated context $h_t$. The high-level view at the top-left shows the information flow in the model aligning with the circled numbers.}
    \label{fig:model}
 \vspace{-4mm}
\end{figure*}

\subsection{Sequence-to-Sequence}
\label{Sequence-to-Sequence}
We model the agent with a LSTM-based sequence-to-sequence architecture \cite{sutskever2014sequence} to control the flow of information, as illustrated in Fig~\ref{fig:model}.
The encoder computes a contextual embedding $\bar{s}_j$ of each token embedding $s_j$ in $S$ by $\bar{s}_{j} = LSTM_{encode}(s_j)$.
At each step $t$ of navigation, the decoder receives the grounded instruction representation $C^*_{t}$ and the aligned image representation $I^*_{t}$ to update its context $h_t$ by $h_{t} = LSTM_{decode}([C^*_{t}, I^*_{t}])$. Finally, we predict the probability distribution of the next navigable viewpoint $p_t$ by $h_t$.
We introduce the method to obtain $C^*_{t}$ and $I^*_{t}$ in Section~\ref{spatial configuration grounding} and Section~\ref{visual grounding}, 
as well as the next viewpoint prediction in Section~\ref{navigable viewpoints prediction}.

\subsection{Spatial Configurations Representation}
To obtain the configurations in a navigation instruction, we first split the instructions into sentences. 
Then we design a parser with rules applied on an off-the-shelf dependency parser\footnote{https://spacy.io/} to extract all the verb phrases
and noun phrases in each sentence. 
In general, each configuration contains at most one motion indicator.
Since we aim to process instructions and look for motions, we split the sentences with the extracted verb phrases as motion indicators to obtain spatial configurations. 
We do not separate the nested configurations with no motion indicator and keep them attached to the dynamic configurations (i.e. the ones with motion-indicator). 
As shown in Figure \ref{fig:spatial Configuration Scheme Example}, "table with chair" is the nested spatial configuration of "move to the table with chair". 
Here, we only consider the prepositions that are attached to verbs, and merge the spatial indicators and motion indicators such as "move to" and use them together as the motion indicator.
After that, we insert a pseudo delimiter token after each configuration and identify their contained noun phrases as landmarks. 
Each navigation instruction $S$ is split into $m$ configurations.
We re-organize the contextual embeddings of tokens  $\left[\bar{s}_1, \bar{s}_2, \cdots\right]$ generated by the encoder into the array of spatial configurations representation ${C}=\left[C_{1},C_{2} \dots C_{m}\right]$, where $m$ is the number of configurations in the instruction.
In the $i$-th configuration representation $C_{i} = \left[c^{i}_{1}, c^{i}_{2} \cdots, c^i_{\mbox{P}}\right]$, the $j$-th element $c^{i}_{j}$ is the contextual embedding of the corresponding $k$-th tokens in the instruction: $c^{i}_{j}=\bar{s}_k$. The last token of each configuration is always the pseudo delimiter indexed by $\mbox{P}$, which contains the most comprehensive context information about the preceding words.
Soft attention is widely used to merge a collection of representations $V$ into one by weighted sum based on the relevance indicated by their associated keys representations $K$ and a query $Q$, calculated by Eq.~\ref{softattention}.
\begin{equation}
\label{softattention}
\mbox{SoftAttn}(Q; K; V) = \mbox{softmax}\left( \frac{Q^{T}WK}{\sqrt{ d_k }} \right) V
\end{equation}
where $W$ is a trainable linear mapping, and $d_k$ is the dimension of each representation in $K$.
We apply a soft attention to each configuration representation with the pseudo delimiter representation $c^{i}_{\mbox{P}}$, which can be calculated by Eq.~\ref{configuration representation}.
\begin{equation}
\label{configuration representation}
\bar{C}_i = \mbox{SoftAttn}_{\mbox{config}}(Q=c^{i}_{\mbox{P}}; K=C_i;V=C_i)
\end{equation}
\noindent After obtaining configuration representations, an agent needs to identify which configuration to follow at each step. To achieve this, we incorporate the intra-configuration and inter-configuration knowledge. Concretely, intra-configuration knowledge is the motion indicator that guides the agent movement and the landmarks that could be 
grounded into the objects in visual images; inter-configuration 
knowledge is that configurations should be processed one after another.

As mentioned above, we identify verbs and noun chunks in configurations as motion indicator and landmarks respectively. 
Each configuration can contain only one motion indicator and multiple landmarks.
Formally, for the $i$-th configuration $C_i$, the motion indicator representation is denoted as $c^i_{\mbox{M}}$ and the landmark representation is denoted as $c^i_L = \left[c^i_{\mbox{L}_1},c^i_{\mbox{L}_2},\cdots,{c^i_{\mbox{L}_p}}\right]$, where $p$ is the number of landmarks. If there is no landmark in the configuration, the value of $c^i_{\mbox{L}}$ will be set as zeros. To enhance the motion indicator and landmark information, we concatenate their word embedding with the configuration representation. In case there are multiple noun chunks in configuration, to simplify, we select the noun closest to the root of the parsing tree as the main landmark, denoted as $\hat{p}$. Then the enriched configuration representation is denoted as
$\tilde{C_i}=\left[\bar{C}_i;c^i_{\mbox{M}};c^i_{\mbox{L}_{\hat{p}}}\right]$.

\subsection{Visual Representation}
\label{visual representation}
To execute a series of configurations, the agent needs to keep track of the sequence of images observed along the navigation trajectory. We firstly transform the low-level image features from ResNet of $n$ navigable images $I=\{I_{1},I_{2},\dots,I_{n}\}$ to $I'=\left[I^{'}_1,I^{'}_2,\cdots,I^{'}_n\right]$ by a fully-connected layer $I^{'}_j=\mbox{FC}_{\mbox{img}}(I_j)$. Then, a soft attention is applied to $I^{'}$ with the previous context $h_{t-1}$, as shown in Eq.~\ref{image softattention}.
\begin{equation}
\label{image softattention}
    \bar{I} = \mbox{SoftAttn}_{\mbox{img}}(Q=h_{t-1};K=I^{'};V=I^{'})
\end{equation}

Furthermore, we equip the agent with object-based representation. Specifically, we get top-K object representations from each image with an object detection model\footnote{We employ Faster R-CNN pre-trained on Visual Genome, and use at most 36 objects that have an area greater than 10 pixels.}.
In this paper, we consider two kinds of object representation: object label representation and object visual representation.
Specifically, the label representation uses the GloVe embedding \cite{pennington2014glove}~of the type of the object, and visual representation uses the region-of-interest~(ROI) pooling of the object detection model. We will compare the two representations and a hybrid representation of them in Appendix~\ref{Representation Analysis}.
Formally, the object representations could be denoted as $O = \left[O_{1}, O_{2} \dots O_{n}\right]$, where for image $I_j$, there is $O_{j} = \left[o_{j,1}, o_{j,2}, \cdots, o_{j,K}\right]$. $o_{j,k}$ is the $k$-th object representation in $j$-th image.

\subsection{Spatial Configuration Grounding}
\label{spatial configuration grounding}
To guarantee the sequential execution, we design a state attention mechanism over the configurations.
We consider the attention weight at each step as a state that measures navigation progress and is updated by a controller.
Formally, the $i$-th configuration at step $t$ is denoted as $\alpha_{t,i}$.
At the first step, the attention weight is initialized to be focused on the first configuration $\alpha_{0}=\left[1,0,\cdots\right]$. 
At each of the following steps, the attention weight is updated by a controller $\gamma_t$ with discrete convolution. $\gamma_t$ is a two dimensional probability distribution indicating to what extent the agent should execute the current configuration or move to the next. The updating process is formally defined in Eq.~\ref{state_aware attention eq}. 
\begin{equation}
\label{state_aware attention eq}
\begin{aligned}
    \alpha_{t,i} &= \sum_{\imath=i-1}^{i} \alpha_{t-1,\imath}\cdot\gamma_{t,i-\imath}
\end{aligned}
\end{equation}
Using a set of rules to determine the value of the controller $\gamma$ is not practical.
For example, for the instruction "move to the table" or "move past the table", it is hard for an agent to decide whether to execute the current configuration or to move to the next one only based observing or not-observing the~"table". To address this issue, we let the agent learn the value of $\gamma$ based on three aspects of information. The first one is the previous hidden state $h_{t-1}$; the second one is the attended image representation $\bar{I_t}$ at the current step; the third one is the similarity score~$S_t$ between the landmark representations and the object representations,
Eq.~\ref{similarity score equation} shows how to get the similarity score~$S_t$, and $\alpha_{t-1}$ is the
attention weight at the previous step. 

\begin{equation}
\label{similarity score equation}
\begin{aligned}
    S_t = \tilde{C}_{L}\cdot O \cdot \alpha_{t-1}
\end{aligned}
\end{equation}
Then, we use a fully connected layer to predict the distribution
$\gamma_{t} = \mbox{FC}_{\gamma}\left(\left[h_{t-1}; \bar{I_t}; S_t\right]\right)$.
Finally, we apply the state attention to $\tilde{C}$ to get the grounded instruction representation based on the configuration
$\hat{C} = \sum_{i} \alpha_{t,i} \cdot \tilde{C_i}$,
which is used as the language input to the decoder $C^*_t = \hat{C}$.

\subsection{Visual Representation Alignment}
\label{visual grounding}
The intuition to leverage the object representation is to select navigable images by aligning the object representation with the configuration representation.
We use two levels of soft attention, first over the objects in each image by configuration representation $\hat{C}$, and second over all images guided by the previous context $h_{t-1}$.
\begin{equation}
    \label{object softattention}
\begin{aligned}
    \hat{O}_j &= \mbox{SoftAttn}_{\mbox{obj}}(Q=\hat{C};K=O_j;V=O_j) \\
    \hat{I} &= \mbox{SoftAttn}_{\mbox{objimg}}(Q=h_{t-1};K=\hat{O};V=I^{'})
\end{aligned}
\end{equation}
where $\hat{O} = \left[\hat{O}_1, \hat{O}_2, \cdots, \hat{O}_n\right]$.
We use the image representation $\hat{I}$, that has aligned the objects with the configurations, as the visual input to the decoder
$I^*_t=\hat{I}$. 

\subsection{Navigable Viewpoint Selection}
\label{navigable viewpoints prediction}
We obtain a new decoder context $h_t$, as described in Section~\ref{Sequence-to-Sequence}, with configuration input $C^*_t$ and visual input $I^*_t$, where $t$ is the current step. The next
step is to predict the viewpoint with the image that has the highest correlation with the current context and configuration, calculated by $z_{t,j} = \left< I^{'}_j, \mbox{FC}_{\mbox{pred}}\left(\left[C^*_{t}; h_t\right]\right) \right>$,
where $\mbox{FC}_{\mbox{pred}}(\cdot)$ is a fully-connected layer.
We sum the scores of the three elevations for each navigable viewpoint $k$ as $\zeta_{t,k}=\sum_{j\in \kappa_{k}} z_{t,j}$, where $\kappa_{k}$ is the set of three elevations' image indexes.
The predicted navigable viewpoint distribution $p_{t}$ can be calculated with $p_{t}=\mbox{softmax}(\zeta_{t})$.

\subsection{Training and Inference}
\label{Training and Inference}
We train our model with two state-of-the-art training strategies in this task.~(1)~\textbf{T1}: We follow Self-Monitor~\cite{ma2019self} optimizing the model with a cross-entropy loss to maximize the
likelihood of the ground-truth navigable viewpoint given by the model,
and a mean squared error loss to minimize the normalized distance in units of length from the current viewpoint to the goal destination. 
At each step, the next viewpoint is selected by sampling the predicted probability of each  navigable viewpoint.~(2)~\textbf{T2}: We follow \cite{tan2019learning} training the model with the mixture of Imitation Learning and Reinforcement Learning, where Imitation Learning minimizes the cross-entropy loss of the prediction and always samples the ground-truth navigable viewpoint at each time step, and Reinforcement Learning uses policy gradient to update the parameters of the model.

During inference, we conduct a greedy search with the highest probability of the next viewpoints to generate the trajectory.
It should be noticed that beam search with a beam size greater than one is not practical because the agent needs to move forward and backward in the physical world, resulting in a long trail trajectory before making a decision.

\section{Experimental Setup}
\begin{table*}[!t]
\small
\renewcommand{\tabcolsep}{0.5em}
    \begin{center}
    \begin{tabular}{c c c c c c c c c c c}
    \hline
      & & \multicolumn{3}{c}{Validation-Seen} & \multicolumn{3}{c}{Validation-Unseen} & \multicolumn{3}{c}{Test(Unseen)}\\
    \hline
        & Method &  NE $\downarrow$ & \textbf{SR} $\uparrow$  & \textbf{SPL} $\uparrow$& NE $\downarrow$& \textbf{SR} $\uparrow$ & \textbf{SPL} $\uparrow$& NE $\downarrow$& \textbf{SR} $\uparrow$& \textbf{SPL} $\uparrow$\\
    
    \hline
    1 & Random~\cite{anderson2018vision} & 9.45 & 0.16 & -&9.23 & 0.16 & - &9.77 & 0.13 & 0.12 \\
    2 & Student-forcing~\cite{anderson2018vision} & 6.01 & 0.39 & - & 7.81 & 0.22 & -& 7.85 & 0.20 & 0.18\\
    3 & Speaker-Follower \cite{fried2018speaker} & 4.36 & 0.54 &  - & 7.22 & 0.27  & - & - & - & - \\
    \hline
    4 & Speaker-Follower* & 3.66 & 0.66 & 0.58 & 6.62 & 0.36 & - &6.62 & 0.35 & 0.28 \\
    5 & Self-Monitor*~\cite{ma2019self} & \textbf{3.22} & \textbf{0.67} & 0.58 & 5.52 & 0.45 & 0.32 & \textbf{5.67} & 0.48 & 0.35 \\

    6 & Environment Dropout*~\cite{tan2019learning} & 4.19 & 0.58 & 0.55 & \textbf{5.43} & \textbf{0.48} & \textbf{0.44} & - & \textbf{0.52} & \textbf{0.47}\\
    7 & Environment Dropout + BERT* & 4.40 & 0.61 & 0.57 & 5.54 & 0.46 & 0.43 & - & - & -  \\
    \hline
    8 & SpC-NAV*& 4.09 & 0.65 & \textbf{0.61} & 5.92 & 0.45 & 0.42 & 6.22 & 0.46 & 0.44 \\
    \hline
    \end{tabular}
    \end{center}
    \vspace{-3mm}
    \caption{\textbf{Experimental Result comparing with baseline models.} * means data augmentation.}
    \label{experiment_table}
       \vspace{-2mm}
\end{table*}

\begin{table}[!t]
\scriptsize
\renewcommand{\tabcolsep}{0.2em}
    \begin{center}
    \begin{tabular}{c c c c c c c c c c}
    \hline
      & \multicolumn{3}{c}{Val-Seen} & \multicolumn{3}{c}{Val-Unseen} & \multicolumn{3}{c}{Test(Unseen)}\\
    \hline
         Method &  NE $\downarrow$ & \textbf{SR} $\uparrow$  & \textbf{SPL} $\uparrow$& NE $\downarrow$& \textbf{SR} $\uparrow$ & \textbf{SPL} $\uparrow$& NE $\downarrow$& \textbf{SR} $\uparrow$& \textbf{SPL} $\uparrow$\\
    
    \hline
    Self-Monitor (T1)  & 3.72 & 0.63 & 0.56 & 5.98 & 0.44 & 0.30 & - & - & - \\
    Sub-Instruction(T1) & - & - & -  & \textbf{6.16} & \textbf{0.42} & \textbf{0.32} & - & - & - \\
    SpC-NAV+T1 & 3.95 & 0.65 & 0.59 & 6.51 & 0.39  & \textbf{0.32} & 6.22 & 0.42 & 0.35\\
    \hline
    EnvDrop (T2) & 4.71 & 0.55 & 0.53 & \textbf{5.49} & \textbf{0.47} & \textbf{0.43} & - & - & -\\
    Sub-Instruction(T2) & - & - & - & 5.67 & 0.47 & \textbf{0.43} & - & - & - \\
    SpC-NAV+T2 & \textbf{4.68} & \textbf{0.59} & \textbf{0.56} & 6.68 & 0.44 & 0.39 & 6.25 & 0.45 & 0.43\\
    \hline
    \end{tabular}
    \caption{\textbf{Experimental Result with Different Training Strategies}. T1 and T2 are two training strategies.}
    \label{experiment_strategy_table}
    \end{center}
    \vspace{-5mm}
\end{table}

\noindent \textbf{Dataset}~We evaluate our model with Room-to-Room (R2R) dataset~\cite{anderson2018vision}, which is built upon the Matterport3D dataset \cite{chang2017matterport3d}. This dataset has 7,189 paths and 21,567 instructions with an average length of 29 words. The whole dataset is divided into training, seen validation, unseen validation, and (unseen) test sets. The seen validation set shares the same visual environments with the training set, while unseen validation and test sets contain different environments.

\noindent \textbf{Evaluation Metrics}~We report three evaluation metrics. (1) Navigation Error (NE): the mean of the shortest path distance between the agent's final position and the goal location.
(2) Success Rate (SR): the percentage of the cases where the predicted final position lays within 3m from the goal location.
(3) Success rate weighted by normalized inverse Path Length (SPL): SPL normalize Success Rate by trajectory length~\cite{anderson2018vision}. SPL is recommended as the primary metric because it
considers both the effectiveness and efficiency of navigation performance.

\subsection{Baseline Models}
We mainly compare Spc-NAV with the following baseline models. \textbf{Seq2Seq}~\cite{anderson2018vision} trained an encoder-decoder model with two learning strategies of random and student-forcing. \textbf{Speaker-Follower}~\cite{fried2018speaker} introduced a speaker module to synthesize new instructions to train the follower module. 
\textbf{Self-Monitor}~\cite{ma2019self} co-grounded instructions and image based on soft attention mechanism.
\textbf{Environmental Dropout} \cite{tan2019learning} proposed a neural agent trained with the method of the mixture of Imitation Learning and Reinforcement Learning. \textbf{Sub-instruction}~\cite{hong2020sub} segmented the instruction into sub-instructions and designed a shifting attention module to ensure the sequential execution order between sub-instructions. The differences between Sub-instruction and our model has been discussed in Section \ref{Related Work}.
\subsection{Implementation Details}
\label{Implementation Details}
We implement SpC-NAV using PyTorch\footnote{https://pytorch.org/}\cite{paszke2017automatic}. We use 768-d BERT-base~\cite{devlin2018bert} (frozen) as the embedding of the raw instruction, and get its 512-d contextual embedding by LSTM.
We encode the representations of the motion indicator and the landmark in each configuration with 300-d GloVe embedding respectively, and concatenate them with the 512-d configuration representation to obtain the enriched configuration representation (1112-d).
We use 300-d GloVe embedding of object label representation to calculate similarity score $S$ with configuration representation. We trained an auto-encoder to map 2048-d object visual representation from Faster R-CNN to 152-d, and use it to obtain the attended object representation $\hat{O}$.
We optimize using ADAM with learning rate $1e-4$ in batches of 64.
We used a rule-based parser to obtain the spatial configuration and spatial semantic elements. This provides some noisy extractions. Appendix~\ref{split error} includes the details about the accuracy of the parser based on our manual annotations of a subset of instructions.

\section{Results and Analysis}
Table~\ref{experiment_table} shows the main performance metrics of our proposed SpC-NAV, compared with the baseline models on seen/unseen validation set and unseen testing set.
To achieve the best result, SpC-NAV is trained with the training strategy T2 (see Section~\ref{Training and Inference}) and the data augmentation proposed in \cite{tan2019learning}.
Our model improves the performance in the seen environment and obtains competitive results in the unseen environment.
Since we use BERT as the input to the encoder while the baseline models use basic word embeddings, we replace the word representations in Environment Dropout with BERT for a fair comparison.
Although the richer language representations help the performance, our model still achieves better results, especially in the seen environments. It indicates that the spatial configuration and spatial elements indeed improve the agent's reasoning ability.

Training strategies are orthogonal to our work, and our model is friendly to the strategies widely used in the literature (T1/T2) (see Section \ref{Training and Inference}). We evaluate SpC-NAV with both T1 and T2 and compare the results with their baseline models as well as Sub-Instruction. We do not apply data augmentation in this setting. As shown in Table~\ref{experiment_strategy_table}, SpC-NAV achieves consistent improvement in the seen environment compared with all the baselines.
In the unseen environment, training with T1, SpC-NAV outperforms Self-monitor (and is even comparable to it with data augmentation) and performs similarly as Sub-Instruction. However, training with T2, our model does not
outperform Environment Dropout and Sub-Instruction in unseen environments. 
We analyze the errors in Section \ref{unseen environment explaination}.
\begin{figure*}[!ht]
\centering
\hfill
\begin{subfigure}[b]{0.35\textwidth}
\includegraphics[height=40pt]{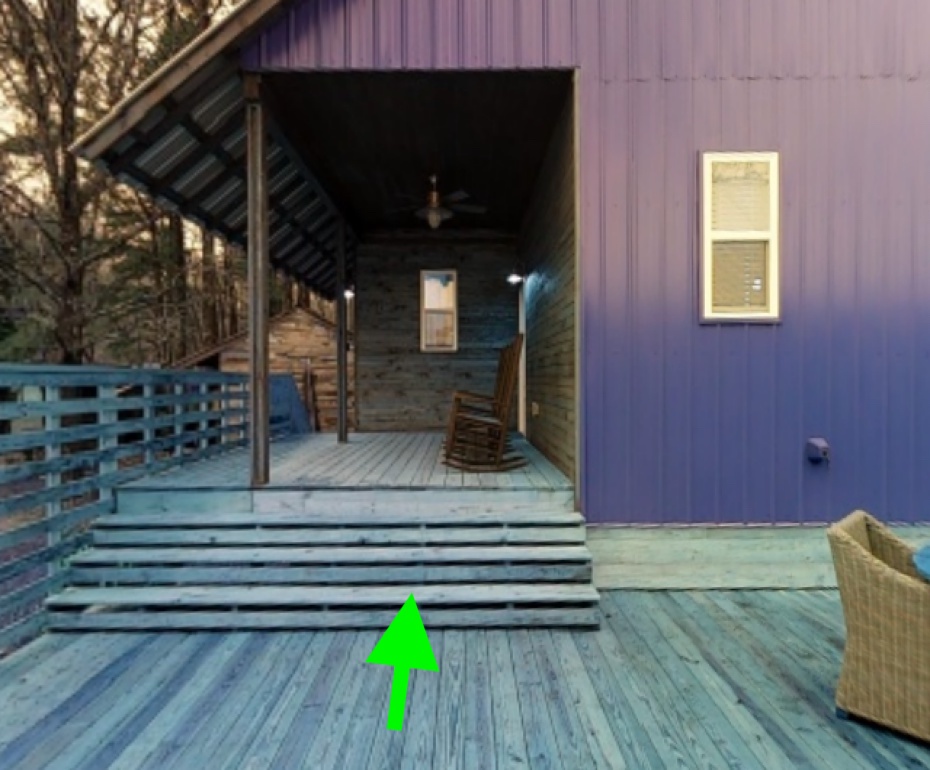}
\includegraphics[height=40pt]{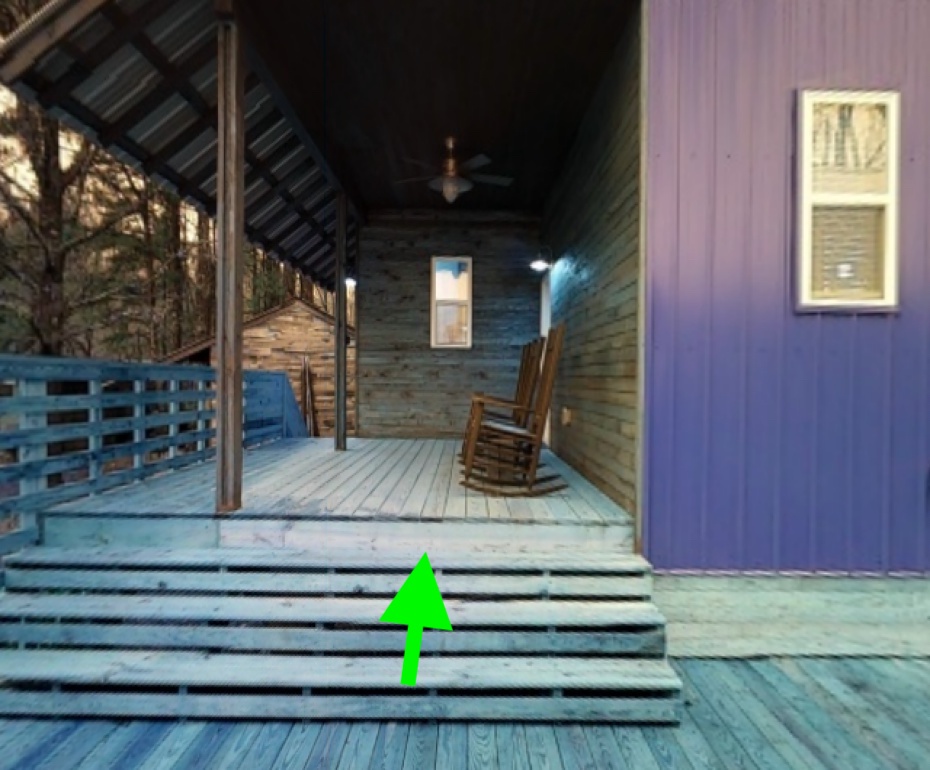}
\includegraphics[height=40pt]{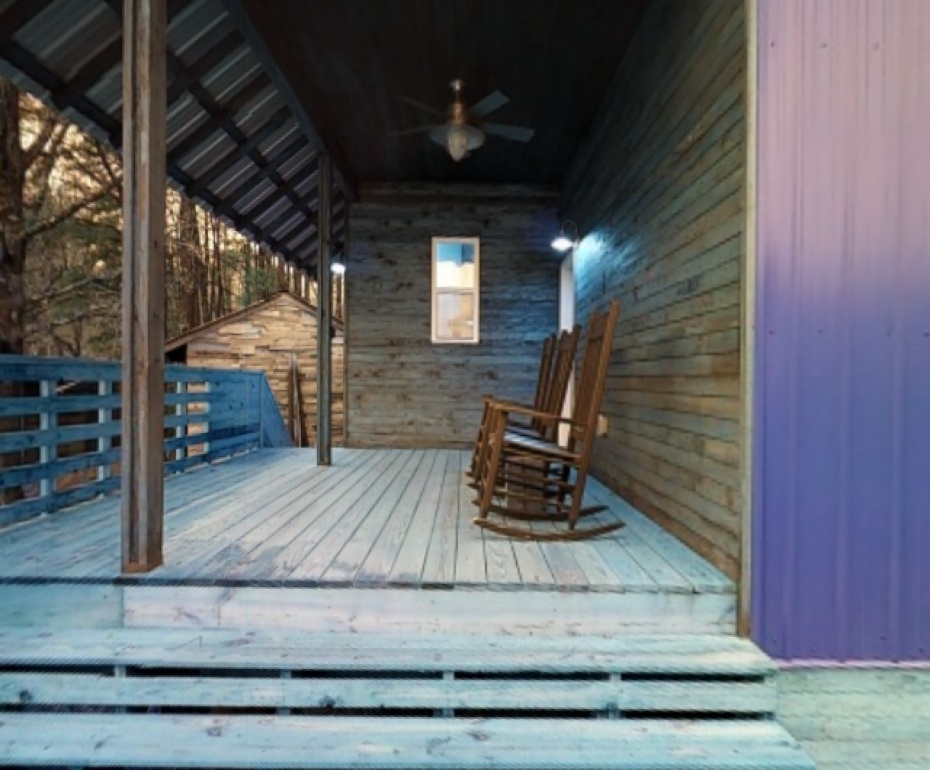}
\\
\includegraphics[height=40pt]{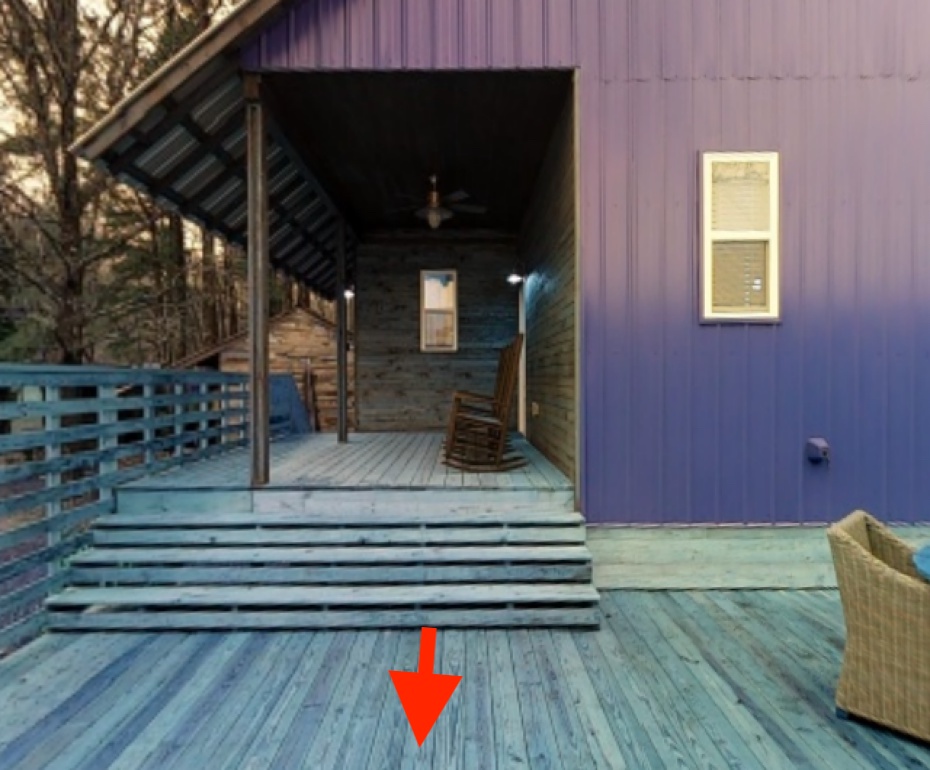}
\includegraphics[height=40pt]{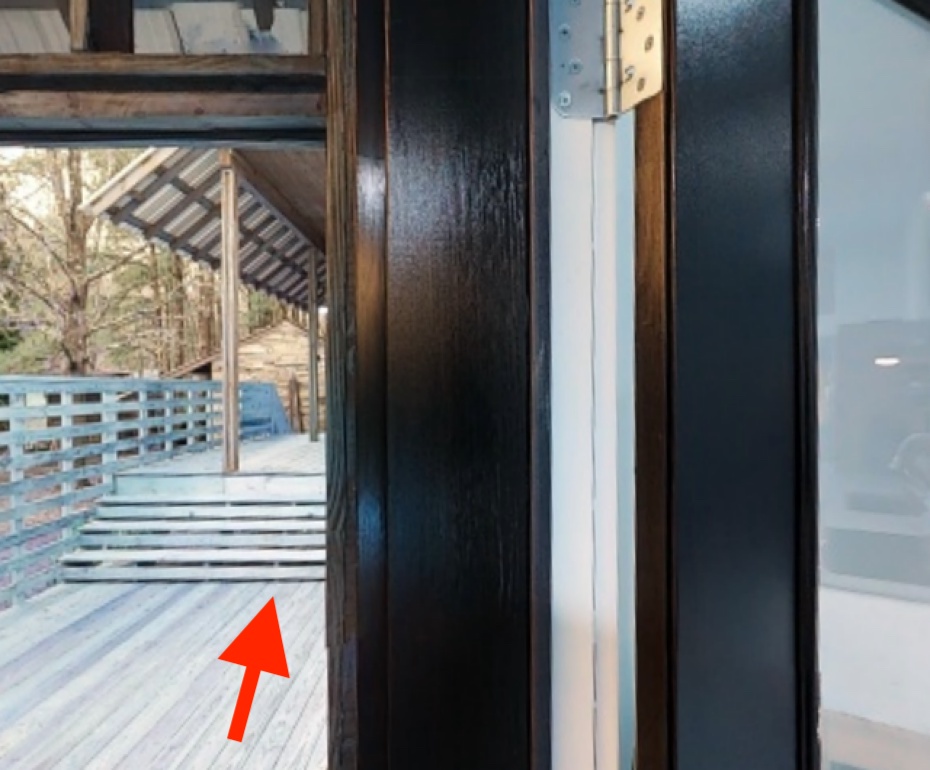}
\includegraphics[height=40pt]{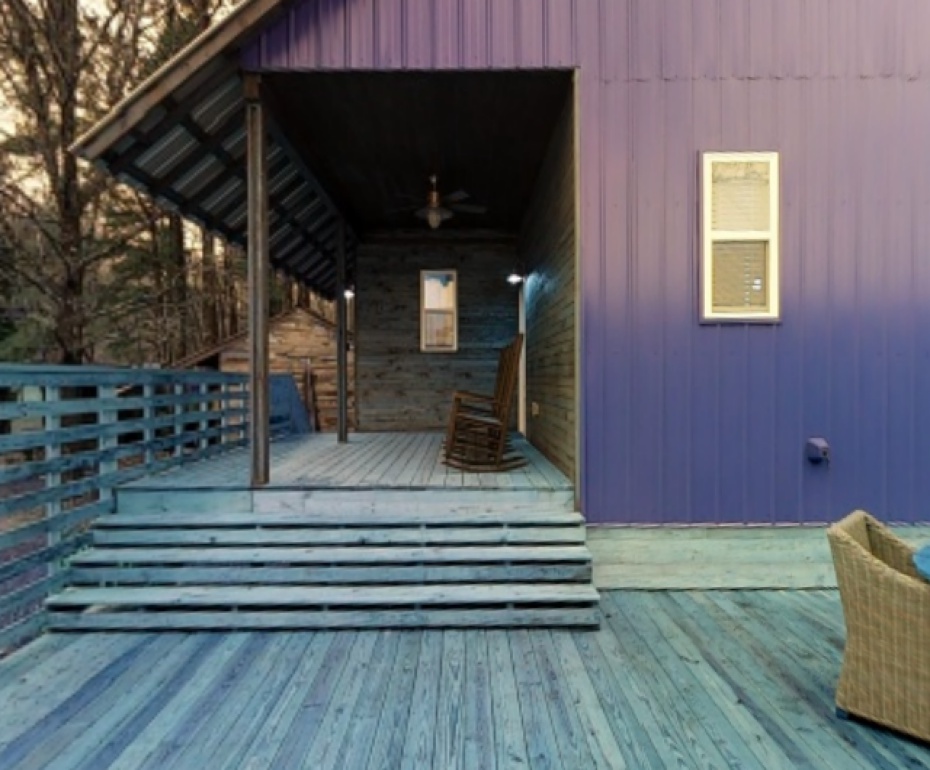}
\caption{$\pm$Motion Indicator
\newline
\scriptsize{Walk up the stairs.\quad\quad\quad\quad\quad}}
\end{subfigure}
\hfill
\begin{subfigure}[b]{0.25\textwidth}
\includegraphics[height=40pt]{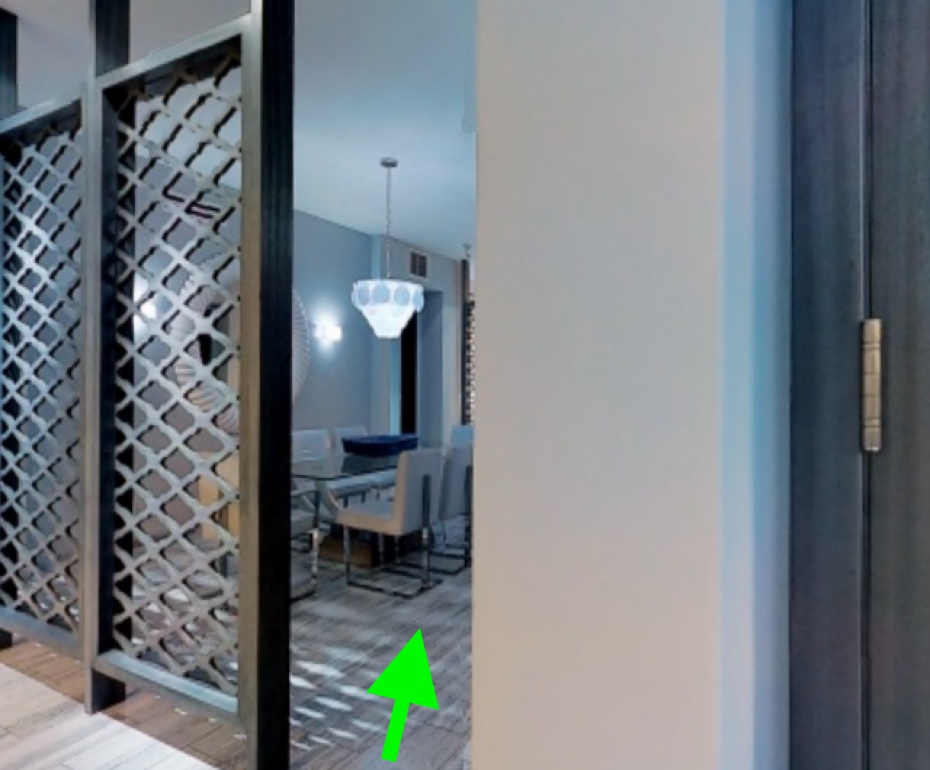}
\includegraphics[height=40pt]{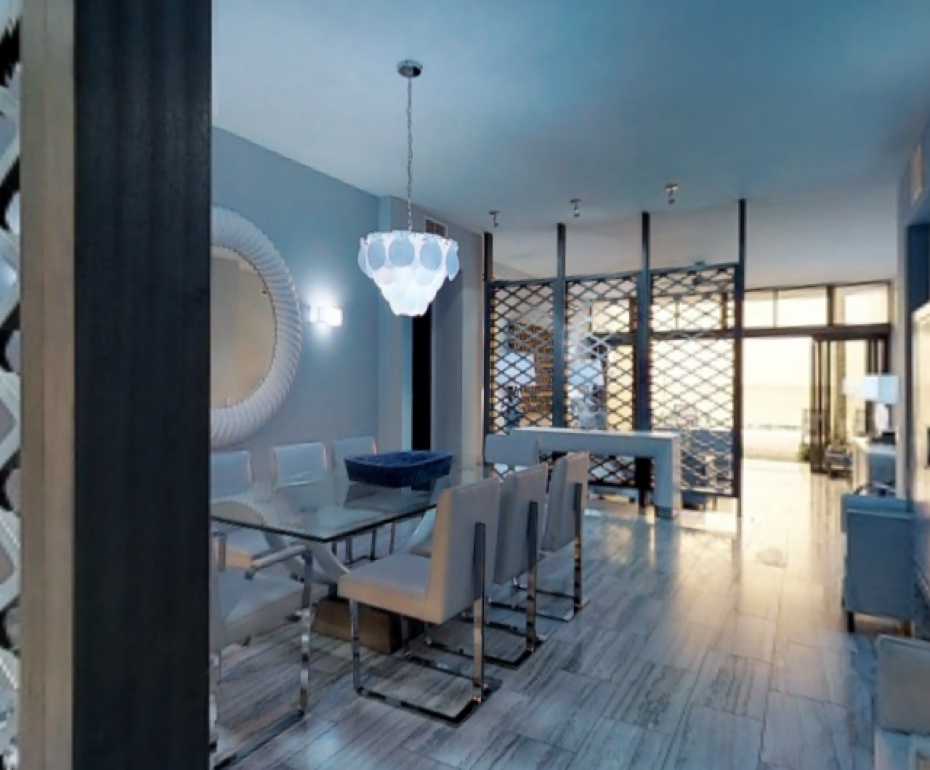}
\\
\includegraphics[height=40pt]{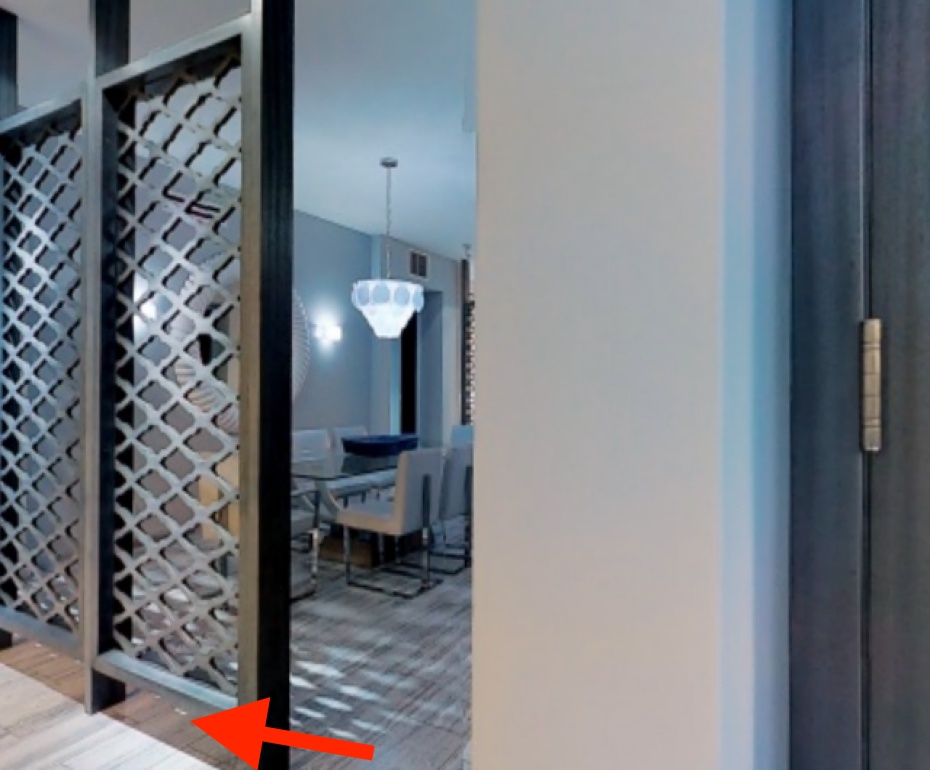}
\includegraphics[height=40pt]{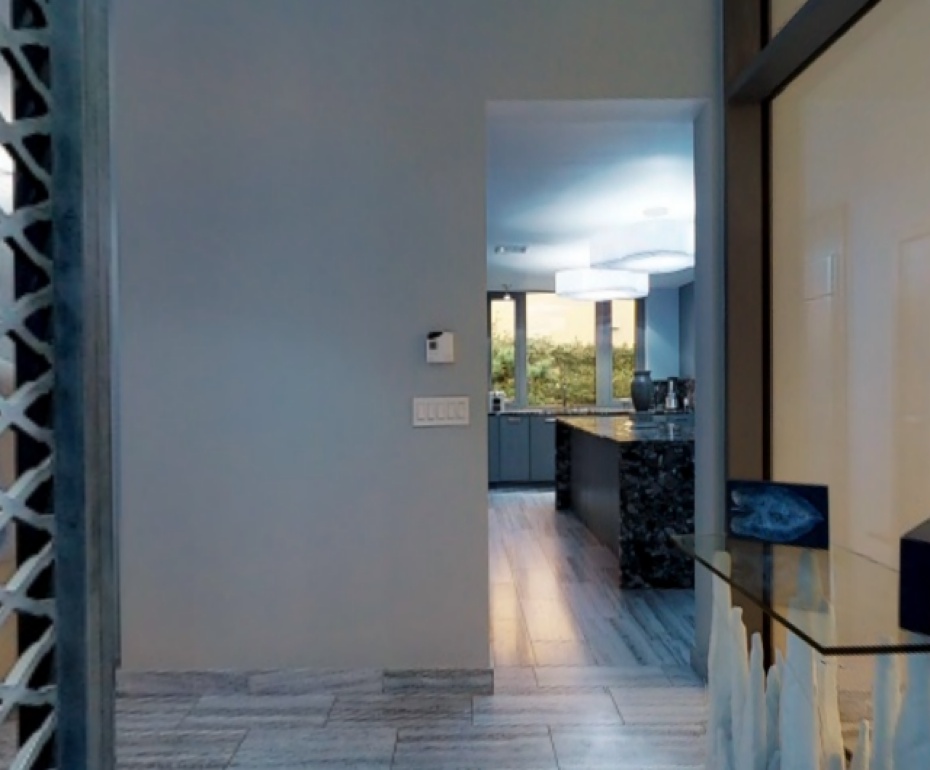}
\caption{$\pm$Landmark
\newline
\scriptsize{Walk past the dinning room table.}}
\end{subfigure}
\hfill
\begin{subfigure}[b]{0.35\textwidth}
\includegraphics[height=40pt]{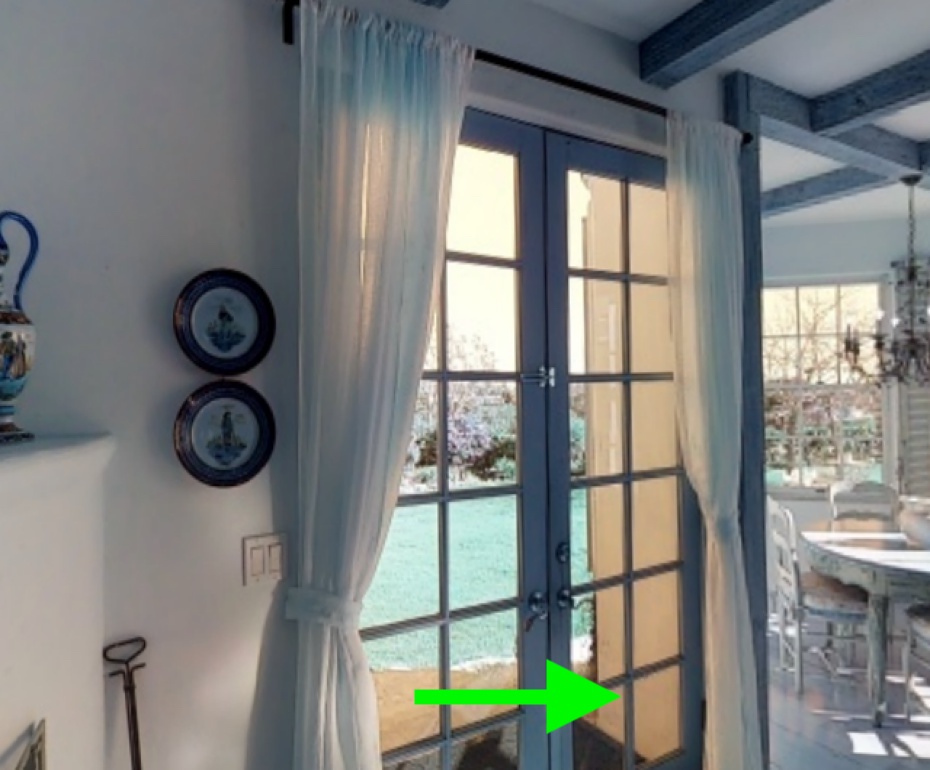}
\includegraphics[height=40pt]{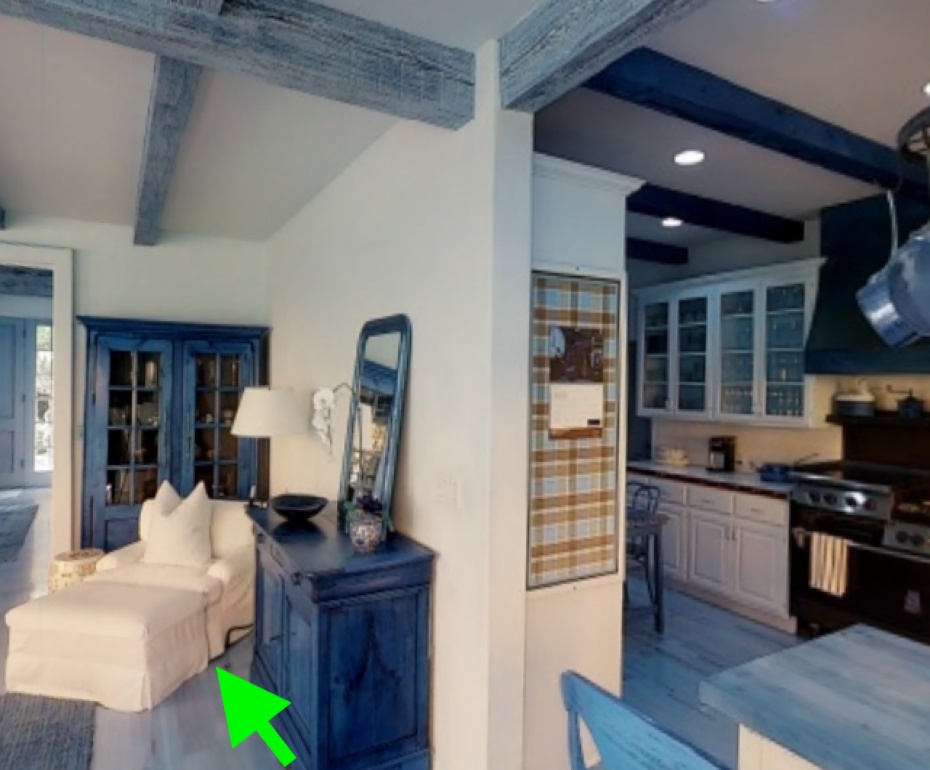}
\includegraphics[height=40pt]{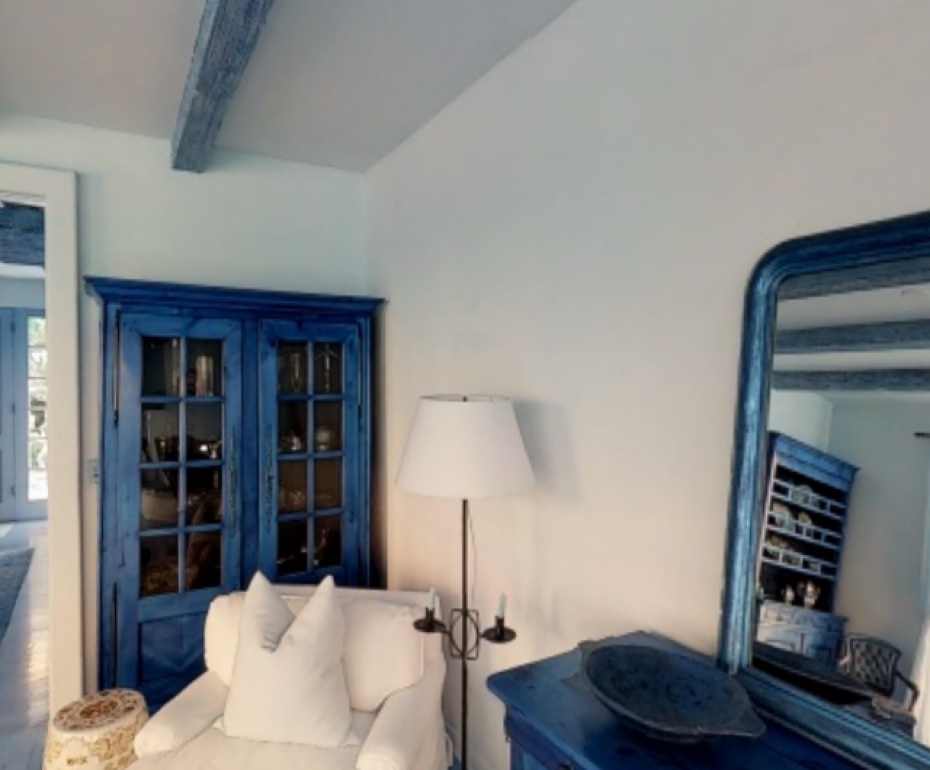}
\\
\includegraphics[height=40pt]{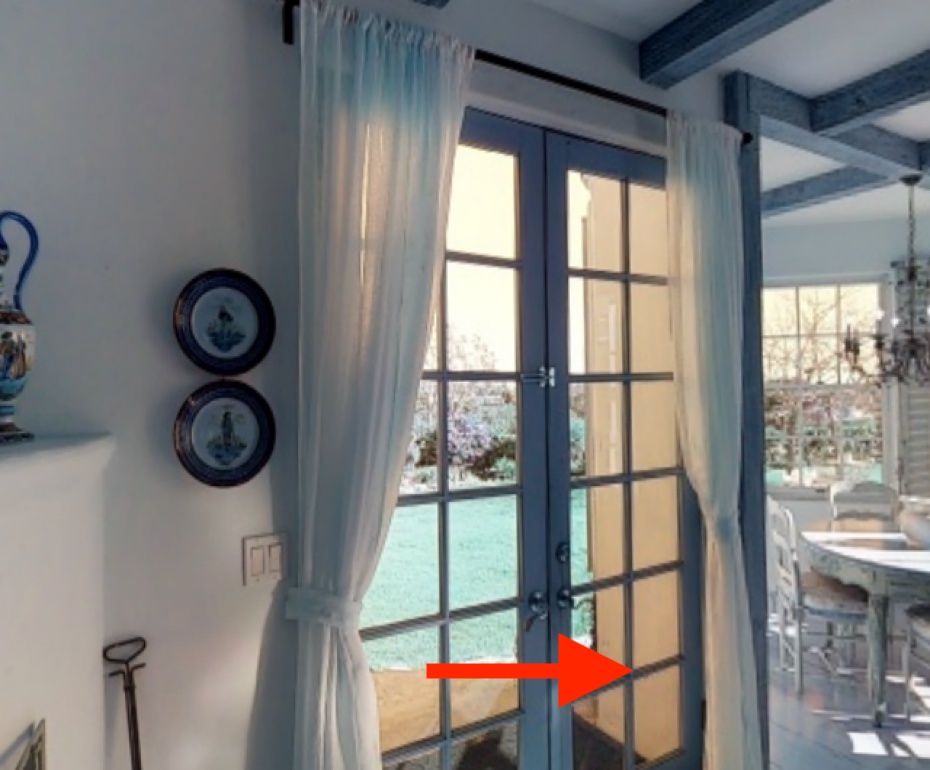}
\includegraphics[height=40pt]{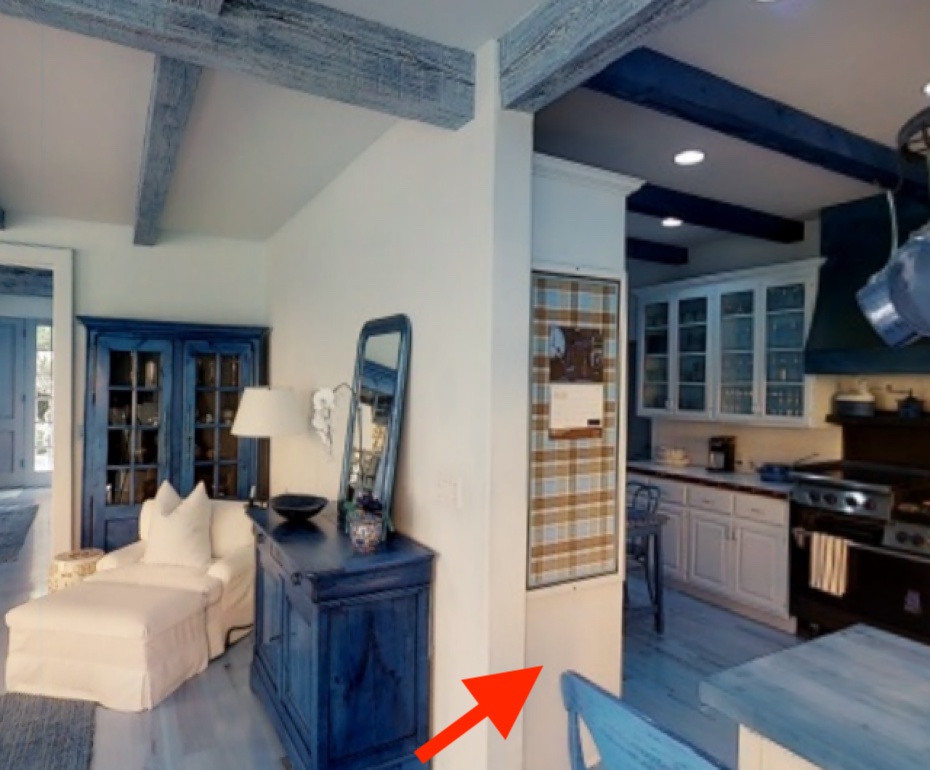}
\includegraphics[height=40pt]{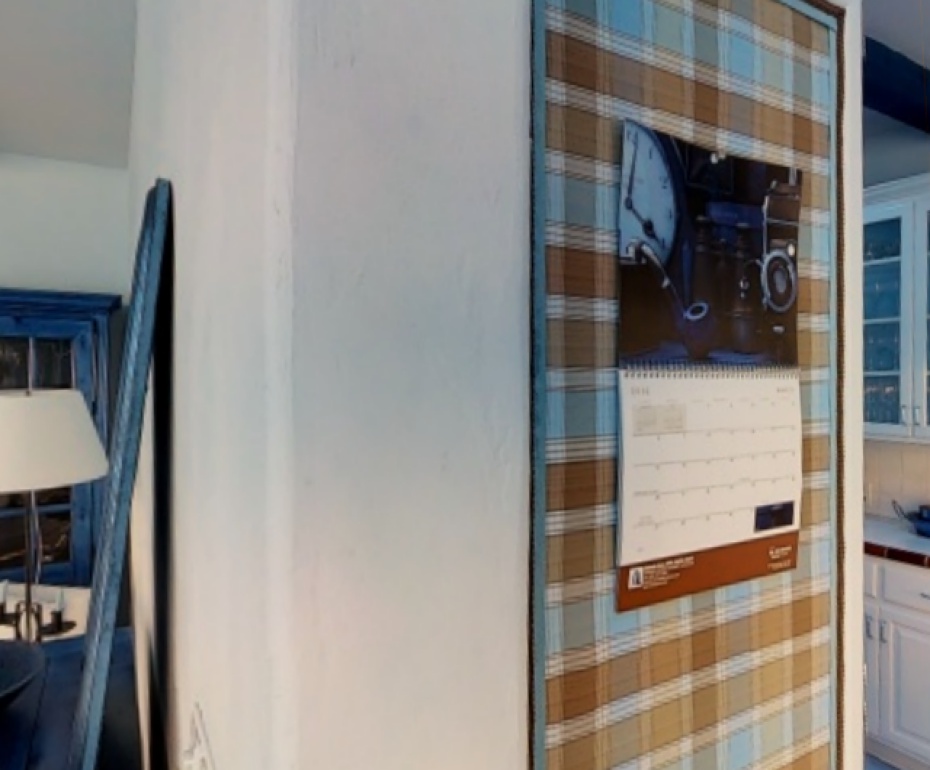}
\caption{$\pm$Similarity Score
\newline
\scriptsize{
Turn right, and walk past the couch.}}
\end{subfigure}
\vspace{-3mm}
\hfill
\caption{\textbf{Analysis of Seen examples.} In these three scenarios, the corresponding spatial configurations are provided. Green arrows in the above figures show the correct trajectory was selected after the additional spatial semantics; red arrows show without that information the agent went wrong.} 
\label{fig: After Adding Motion Indicator}
\vspace{-3mm}
\end{figure*}

\subsection{Ablation Analysis}
Table~\ref{Ablation Study} shows how various spatial semantic elements influence the performance of the model. The model is trained with the training strategy T1. Row\#1 is our model without considering spatial elements.
From row\#2 to row\#3, we incorporate the representations of the motion indicator and the landmark into spatial configuration representation incrementally. In row\#4, we use the similarity score between the landmark representations in the configuration and the object label representations in the image to control the transitions between spatial configurations. All motion indicator, landmark and similarity score improve the performance. After applying the similarity score, the large gain indicates that the connection between landmarks and objects is important in language grounding.
\begin{table}[!ht]
\small
\centering
\renewcommand{\tabcolsep}{0.2em}
\begin{subtable}[t]{0.44\textwidth}
     \begin{tabular}{l l c c c c c c}
    \hline
    & & \multicolumn{3}{c}{Validation-Seen} & \multicolumn{3}{c}{Validation-Unseen} \\
    \hline
     & Model & NE$\downarrow$ & \textbf{SR}$\uparrow$  & \textbf{SPL}$\uparrow$ & NE$\downarrow$ & \textbf{SR}$\uparrow$ & \textbf{SPL}$\uparrow$\\
    \hline
    1 &SpC-NAV & 4.11 & 0.62 & 0.53 & 6.49 & 0.39 & 0.29 \\
    2 &SpC-NAV\textsubscript{\small{M}} & \textbf{3.88} & 0.62 & 0.53 & \textbf{6.21} & \textbf{0.40} & 0.28 \\
    3 &SpC-NAV\textsubscript{\small{M+L}} & 4.01 & 0.62 & 0.54 & 6.27 & 0.39 & 0.29 \\
    4 &SpC-NAV\textsubscript{\small{M+L+S}} & 3.95 & \textbf{0.65} & \textbf{0.59} & 6.51 & 0.39 & \textbf{0.32}\\
    \hline
    \end{tabular}
\end{subtable}
\caption{\textbf{Ablation study with different spatial semantics.} The subscription letters mean the model took those information into account; \textit{M: motion indicator}; \textit{L: landmark}; \textit{S: similarity score}.}
\label{Ablation Study}
\vspace{-6mm}
\end{table}

\subsection{Qualitative Analysis}
\subsubsection*{Seen Environment}
\label{Seen Environment Example}
We analyze some  qualitative examples to find out how the spatial semantics improve the model.
For the semantics of motion, we find that our model can improve the cases that motions contain "up" and "down" after adding the representation of motion indicator.
Figure~\ref{fig: After Adding Motion Indicator}~(a) shows an example of such a scenario. The spatial configuration is "walk up the stairs", and the agent could find the right viewpoints after we incorporated the representation of the motion indicator "walk up". 
However, the model makes more mistakes in the cases that the motion indicators are highly related to the objects, such as "walk through", "walk past", and "walk towards", which need the landmark information. 
In these latter cases, the model should consider both motions and landmarks together.
In another experiment, we added the landmark representation.
Figure \ref{fig: After Adding Motion Indicator}~(b) shows an example that the spatial configurations is "walk past the dinning room table".  The agent can select the correct viewpoints when we incorporate the representation of landmark "dinning room table".
We also analyze the influence of the similarity score, and found that when the information in the current configuration is not sufficient to make a decision, the similarity score will assist in choosing the next configuration. For example, in Figure~\ref{fig: After Adding Motion Indicator}~(c), the spatial configurations are "turn right" and "walk past the couch". Without using the similarity score in controlling the transitions between configurations, the agent tends to select a viewpoint in the "right" direction. But with similarity score, the agent will consider both "turn right" and "walk past the couch", and selects the correct viewpoint that the "couch" can be seen.

\subsubsection*{Unseen Environment}
\label{unseen environment explaination}
\begin{figure}[!t]
\centering
\begin{minipage}[c]{0.28\columnwidth}
\includegraphics[height=45pt]{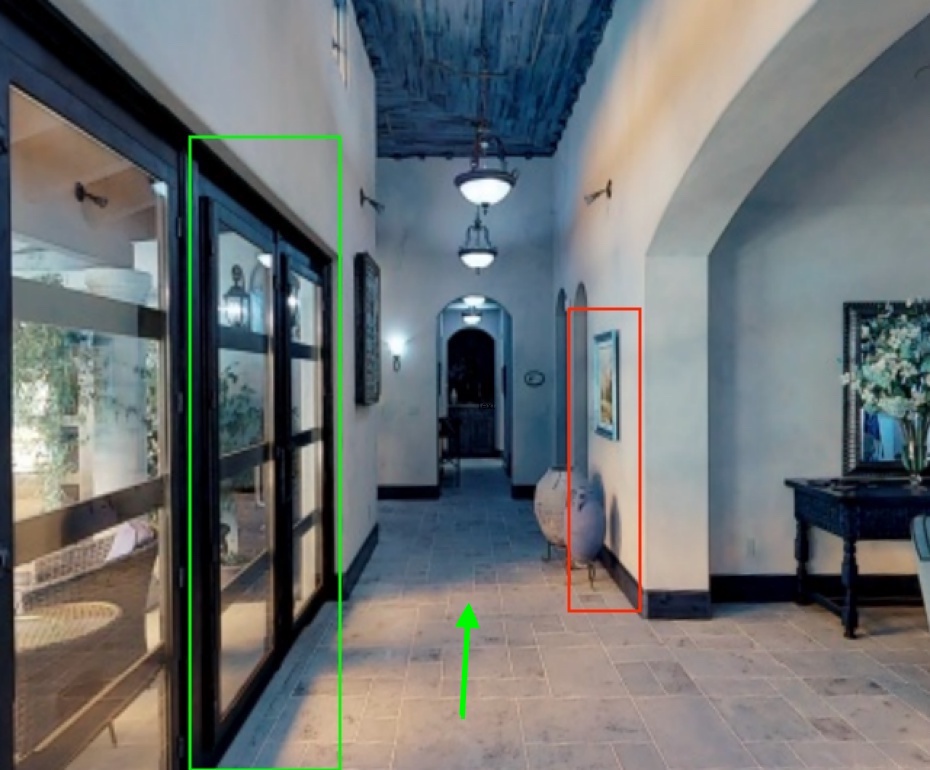}
\end{minipage}
\begin{minipage}[c]{0.28\columnwidth}
\includegraphics[height=45pt]{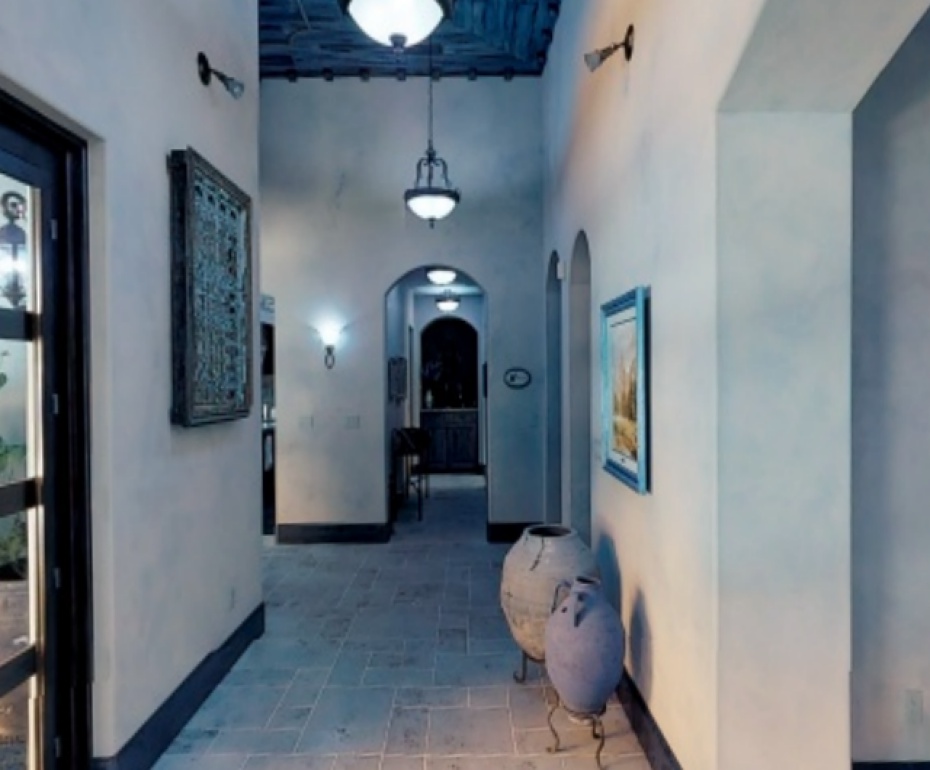}
\end{minipage}
\begin{minipage}[c]{0.35\columnwidth}
\scriptsize{(a) Walk to the \textbf{glass door}.}
\end{minipage}
\\
\begin{minipage}[c]{0.28\columnwidth}
\includegraphics[height=45pt]{unseen_example1_1.jpg}
\end{minipage}
\begin{minipage}[c]{0.28\columnwidth}
\includegraphics[height=45pt]{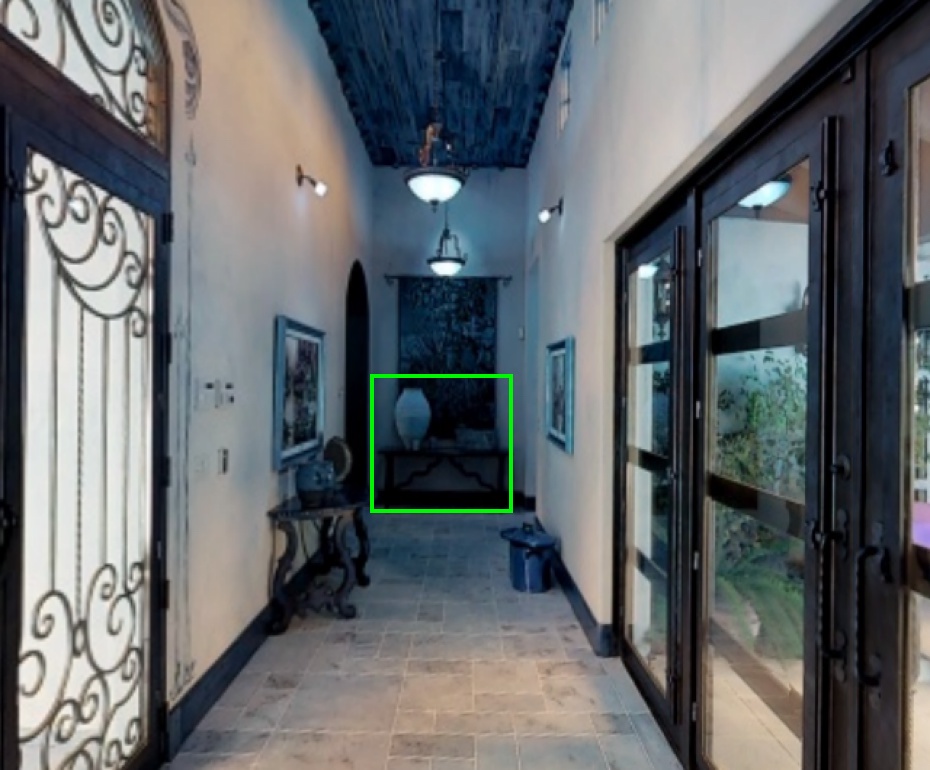}
\end{minipage}
\begin{minipage}[c]{0.35\columnwidth}
\scriptsize{(b) Go to the \textbf{pottery}.}
\end{minipage}
\caption{\textbf{Analysis of an Unseen example}} 
\label{unseen environment}
\vspace{-6mm}
\end{figure}

Table~\ref{experiment_table} and Table~\ref{experiment_strategy_table} show that our model does not outperform Environment Dropout in the unseen environments. We noticed that the main error is that some objects can not be detected in the image by the object detection model. This is more problematic for our model because we explicitly align the landmark phrases with the detected objects.
For example, 
in Fig~\ref{unseen environment}~(a), the agent selects the correct viewpoint when the configuration is "Walk to the glass door" because the connection between the landmark "glass door" and the object "door" has been learned in training set. In Fig (b), the agent is wrong when the configuration is "Walk to the pottery." because the "pottery" is not detected at the initial perspective and the word "pottery" never appears in the training set. However, the agent selects a viewpoint that a bounding box contains a pottery.
The gap between seen and unseen become larger after data augmentation since our model is able to capture the structure of the language by observing more examples. It can deal with the variations in the instructions and improve the performance in the seen environment, but it fails to deal with the novel objects and visual variations in the unseen environments. This is an orthogonal issue addressed in zero-shot learning~\cite{blukis2020few}.
\subsection{State Attention Visualization}
\begin{figure}[!ht]
    {
    \subcaptionbox{\tiny{State Seen}\label{fig:state-seen}}
    {\includegraphics[height=.55in]{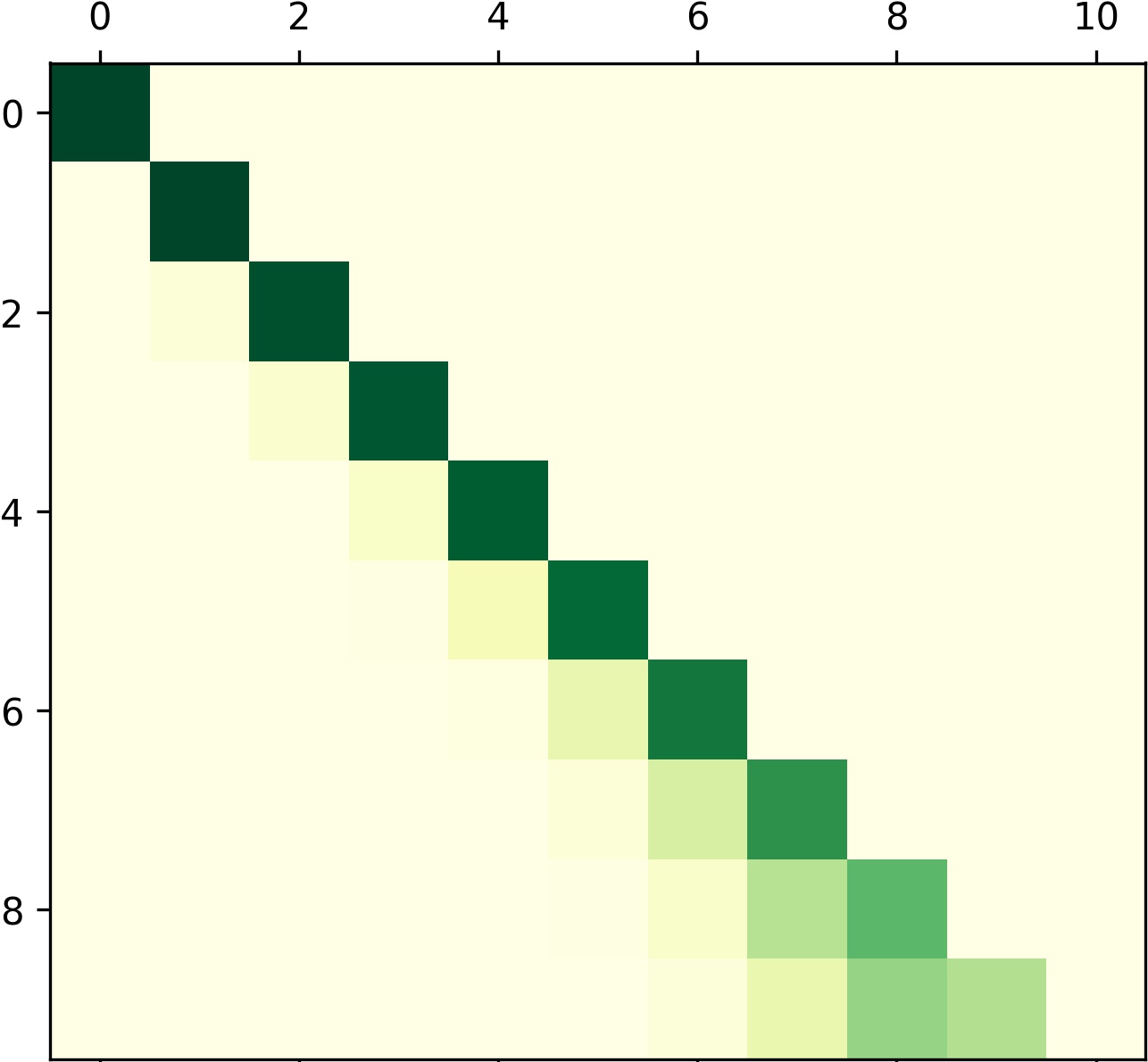}}
    \subcaptionbox{\tiny{Soft Seen}\label{fig:soft-seen}}
    {\includegraphics[height=.55in]{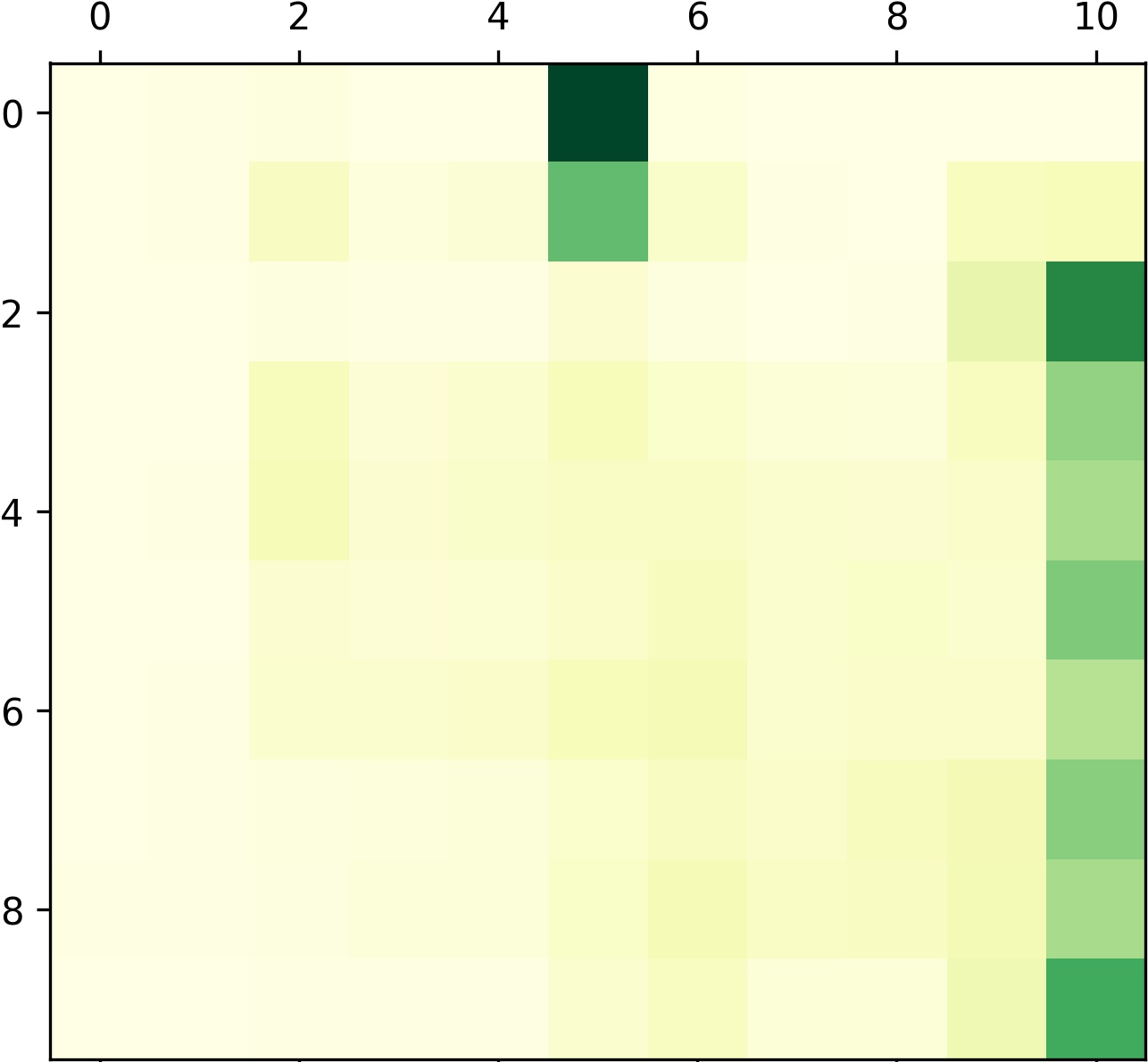}}
    \subcaptionbox{\tiny{State Unseen}\label{fig:state-unseen}}
    {\includegraphics[height=.55in]{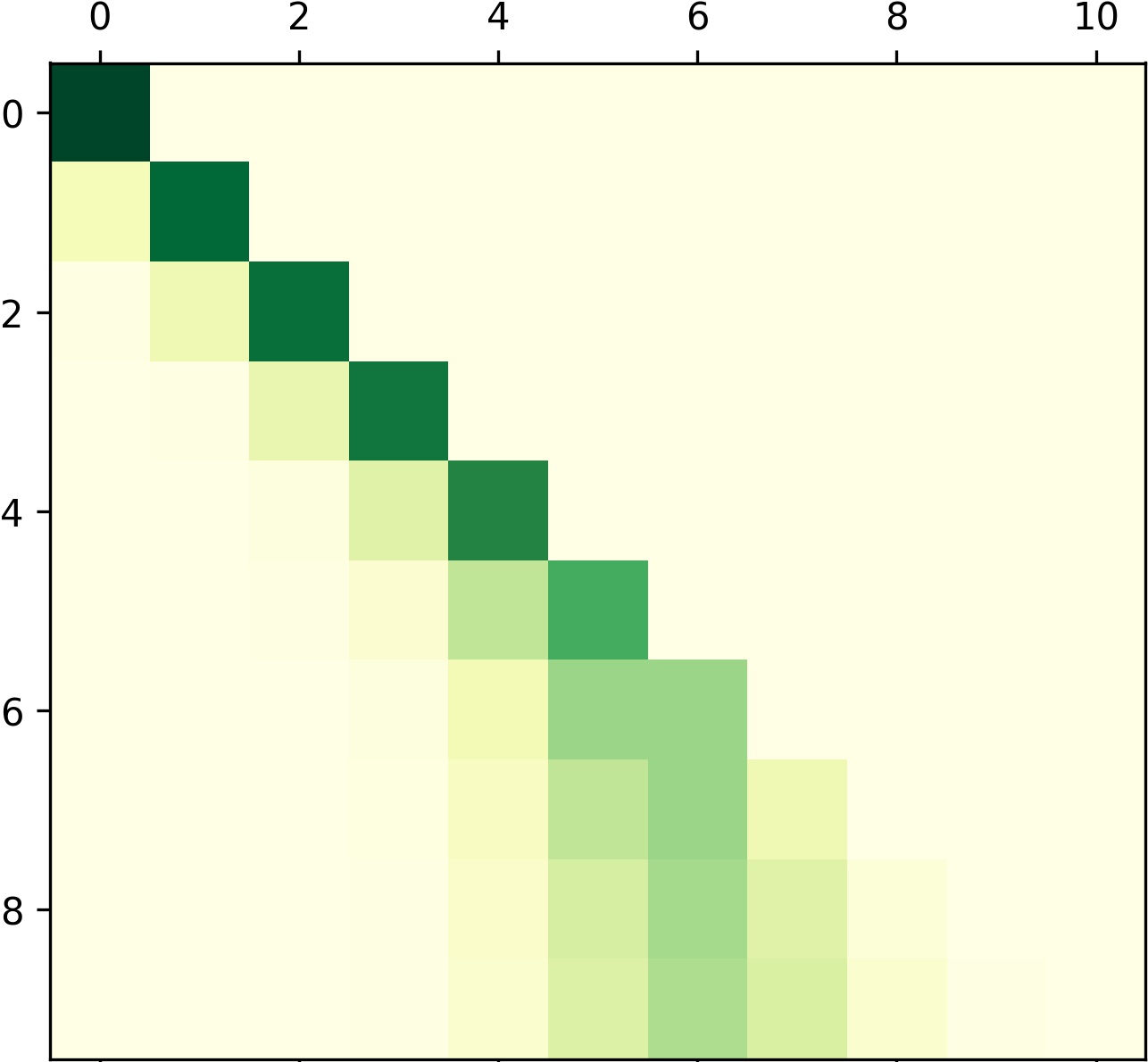}}
    \subcaptionbox{\tiny{Soft Unseen}\label{fig:soft-unseen}}
    {\includegraphics[height=.55in]{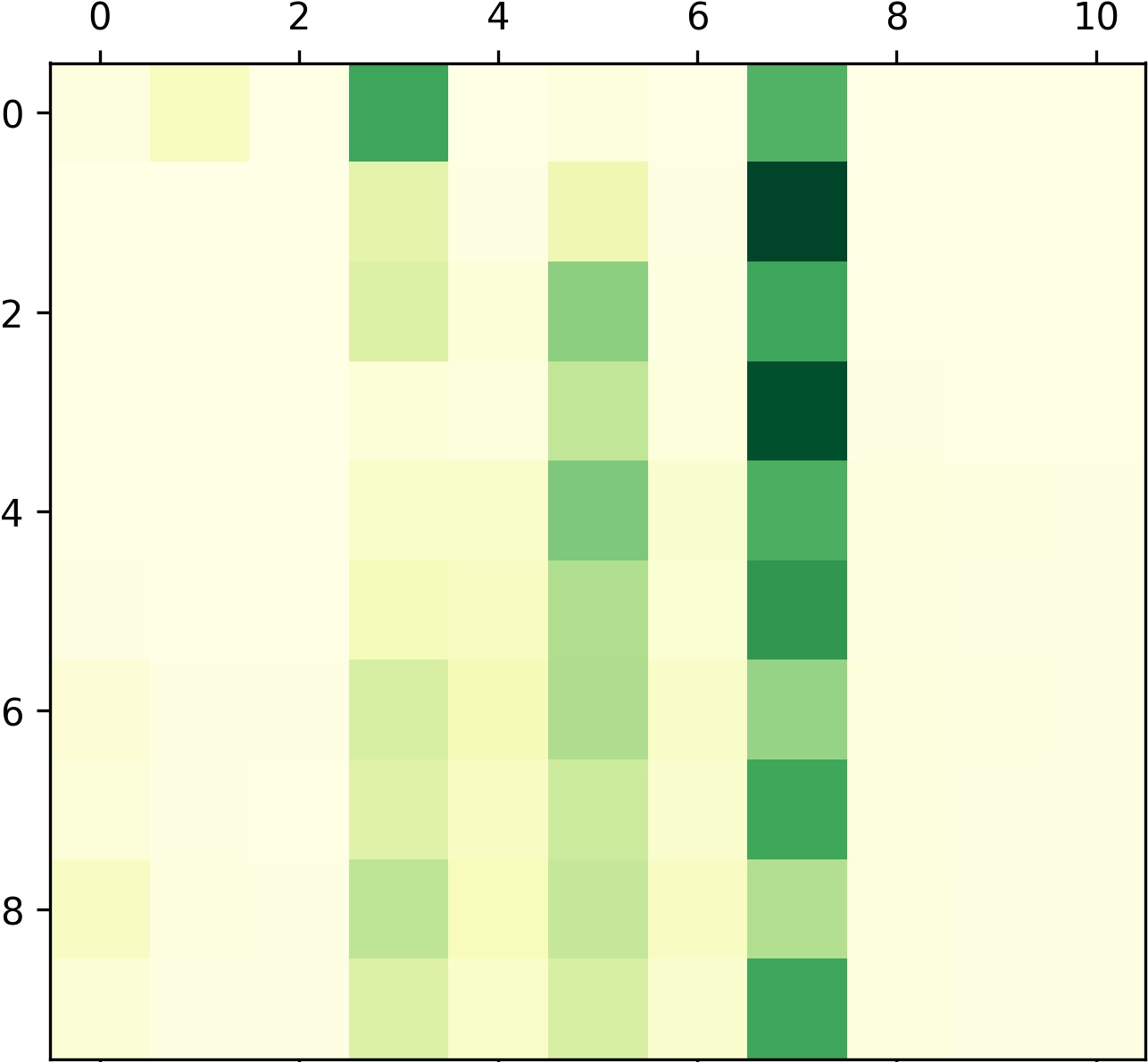}}
    \\
    \subcaptionbox{\tiny{Soft seen of Self-Monitor}\label{fig:80-seen}}
    {\includegraphics[height=.55in]{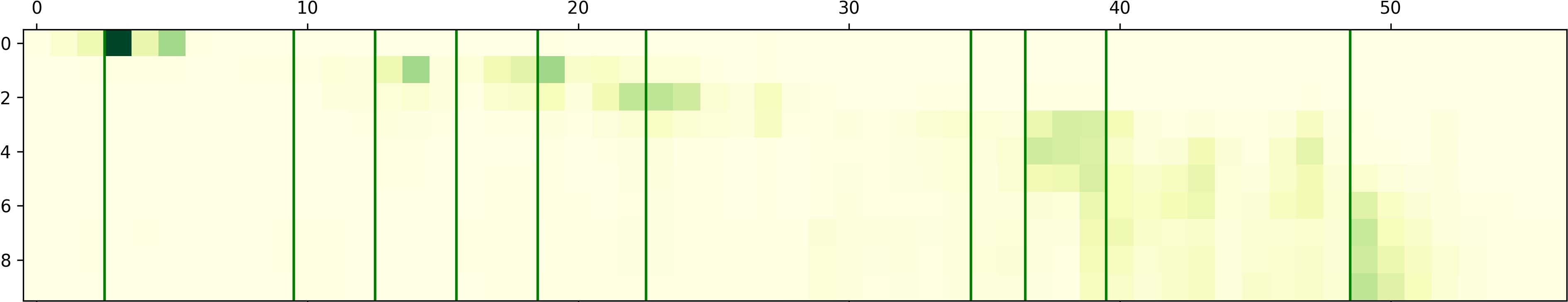}}
    \\
    \subcaptionbox{\tiny{Soft unseen of Self-Monitor}\label{fig:80-unseen}}
    {\includegraphics[height=.55in]{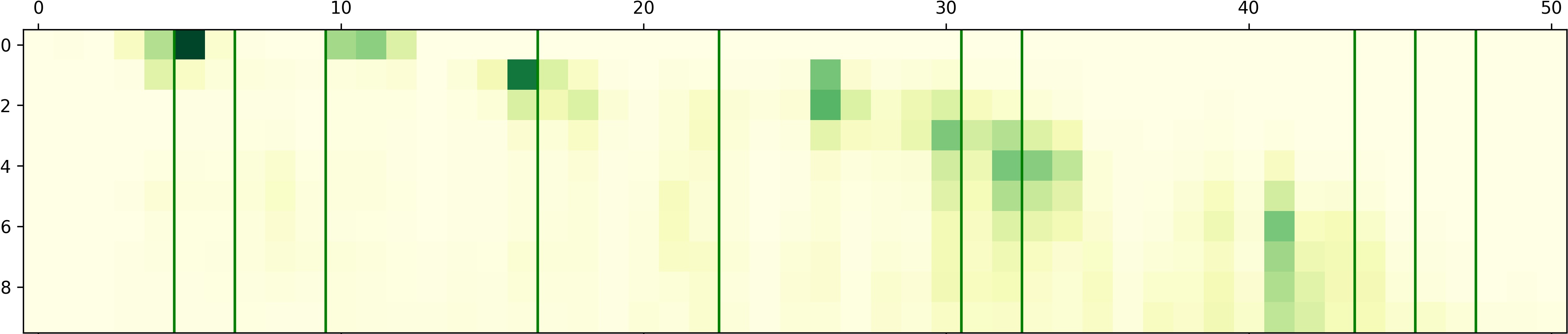}}
    \\
    }
    \vspace{-3mm}
    \caption{\textbf{Attention weights of various attention strategies with seen and unseen examples.} The horizontal axis is the configuration order, and the vertical axis is  the temporal order of the steps taken by the agent. 
    Each row in sub-figures show the attention distribution over the configurations (or tokens) in an instruction at each time step.
    The green vertical lines in Figure (\subref{fig:80-seen}) and Figure (\subref{fig:80-unseen}) indicate the split points of the configurations in the instruction.
    \label{fig:visualization}}
\end{figure}

We visualize the state attention and the soft attention weights over configurations. 
As shown in Fig~\ref{fig:state-seen} and Fig~\ref{fig:state-unseen}, our designed state attention demonstrates that the grounded configuration shifts gradually from the first configuration to the last in both seen and unseen environments.
We apply the soft attention used in Self-Monitor on spatial configurations, as shown in Fig~\ref{fig:soft-seen} and Fig~\ref{fig:soft-unseen}, it can not preserve the sequential execution order.
We also show the soft attention weights of the grounded instruction in the Self-Monitor by splitting the instructions with the boundaries of our configurations.
As shown in Fig~\ref{fig:80-seen} and Fig~\ref{fig:80-unseen}, 
although their attention weights show the gradual shift, 
many configurations are skipped.

\section{Conclusion}

We propose a neural agent that incorporates the semantic elements of spatial language for vision-and-language navigation. 
We use the notion of spatial configurations as the main linguistic unit of the instructions and enhance the spatial configuration representation with the representations of motion indicator and landmark. We design a state attention to guarantee the sequential execution order of configurations and use the similarity score between the representations of landmarks and objects to control the transitions between configurations.
Based on our results, incorporating the spatial semantics improves reasoning ability over navigation. Future work could investigate more fine-grained spatial semantics and the geometry of spatial relations.
Also, we will deal with novel objects in a zero-shot setting to improve the unseen environments results.

\bibliographystyle{acl_natbib}
\bibliography{anthology,acl2021}

\clearpage
\appendix
\section{Appendix}
\label{sec:appendix}

\subsection{Visual Representation Analysis}
\label{Representation Analysis}
In this section, 
we experiment with three types of object representations introduced in Section~\ref{visual grounding}, which are object label representation and object visual representation and the combination of these two types of object representation. 
As shown in Table~\ref{tab:Different VIsual Features COmaprision}, object visual representation performs better in unseen environments, and we use it to get attended object representation $\hat{O}$ in our best model. This experiment does not consider the similarity score between the representations of landmarks and objects.

\begin{table}[!ht]
\renewcommand{\tabcolsep}{0.15em}
\small
\centering
\begin{subtable}[t]{0.44\textwidth}
    \begin{tabular}{c |c c c |c c c}
    \hline
     & \multicolumn{3}{c}{Validation-Seen} & \multicolumn{3}{c}{Validation-Unseen}\\
    \hline
     Repr.  &  NE$\downarrow$ & \textbf{SR}$\uparrow$  & \textbf{SPL}$\uparrow$ & NE$\downarrow$ & \textbf{SR}$\uparrow$ & \textbf{SPL}$\uparrow$ \\
    \hline
    Label & 4.51 & 0.58 & 0.52 & 6.43 & 0.37 & 0.28 \\
    Visual & 4.01 & 0.62 & 0.54 & 6.27 & 0.39 & 0.29\\
    Label + Visual& 4.45 & 0.59 & 0.53 & 6.54 & 0.37 &0.28 \\
    \hline
    \end{tabular}
\end{subtable}
\hfill

\caption{Result with Different Visual Representations.}
\label{tab:Different VIsual Features COmaprision}
\end{table}

\subsection{Parsing Analysis}
\label{split error}
The performance of our rule-based parser influences the result of navigation.
To evaluate it, we manually annotated 845 spatial configurations for 200 instructions. We annotated motion indicators, spatial indicators and landmarks in those configurations.
Our parser achieves an accuracy of 85\% in extracting the spatial configurations. For the extraction of spatial elements, the accuracy is 73\% for motion/spatial indicators, and 77\% for landmarks. 

In the following, we analyze two types of error in getting spatial configurations (Split Error and Order Error), and other errors that generated in the extraction of motion indicator, spatial indicator and landmark.
\subsection*{Split Error} The split configuration may only convey the spatial position of objects rather than executable navigation information. For example, in the instruction, ``Turn left. There is a rocking chair in it,'' two configurations are generated based on our split method: ``Turn left'' and ``There is a rocking chair in it.'' However, the second configuration is not an independent spatial configuration because it indicates no motion, and it is attached to the previous configuration. 

\subsection*{Order Error} We order the configurations based on their occurrence in the sentence. However, there are cases that the configurations have an inverted order. For instance, ``Stop once you pass the counter on the right'' is split as ``stop'' and ``you pass the counter on the right.'' However, the implied sequence is inverted because of ``once''. 

\subsection*{Motion Indicator and Spatial Indicator} We build a vocabulary based on training data to collect the commonly used verb phrases, and the vocabulary size is 241. Table \ref{tab:motion indicator examples} shows some examples. If the motion indicator and spatial indicator does not show in the vocabulary, we will treat the verbs as the motion indicators and prepositions as spatial indicators in configurations.
With this method, we can get 73\% accuracy since there are expressions that never appear in the training dataset, and it is hard to extract the complete verb phrases only based on pos-tag. 

\subsection*{Landmark}
We extract the noun phrases of each configuration as landmark and can get 77\% accuracy.  However, there are some special cases, for example, "a left" in "make a left" is extracted as noun chunk, but it can not be treated as a landmark.
Also, for the expression "middle of the doorway", "the middle" and "the doorway" are both noun chunks, but the whole phrase is the landmark instead of separated ones.

\begin{table}[ht]
\small
\centering
\begin{subtable}[t]{0.44\textwidth}
    \begin{tabular}{|c|}
    \hline
    head straight, walk through, walk down, walk into, \\
    walk inside, turn around, turn left, make a left turn, \\
    jump over, move forward, turn slightly right \\
    \hline
    \end{tabular}
\end{subtable}
\hfill
\caption{Verb Phrases Examples}
\label{tab:motion indicator examples}
\end{table}

\end{document}